\documentclass[11pt]{article}

\usepackage[final]{acl}

\usepackage{times}
\usepackage{latexsym}

\usepackage[T1]{fontenc}

\usepackage[utf8]{inputenc}

\usepackage{microtype}

\usepackage{inconsolata}

\usepackage{graphicx}

\usepackage{algorithm}
\usepackage{algorithmic}

\usepackage{amsmath}
\usepackage{amssymb}

\usepackage{booktabs}
\usepackage{multirow}

\usepackage{subcaption}

\usepackage{tcolorbox}
\tcbuselibrary{listings,breakable}

\usepackage{enumitem}
\newcommand\blfootnote[1]{\begingroup\renewcommand\thefootnote{}\footnote{#1}\addtocounter{footnote}{-1}\endgroup}

\title{\textsc{InfiniteWeb}: Scalable Web Environment Synthesis for GUI Agent Training}

\author{Ziyun Zhang\textsuperscript{1*} \quad Zezhou Wang\textsuperscript{2*}  \quad Xiaoyi Zhang\textsuperscript{3$\dagger$} \\
\textbf{Zongyu Guo$^{3}$} \quad \textbf{Jiahao Li$^{3}$} \quad \textbf{Bin Li$^{3}$} \quad \textbf{Yan Lu$^{3}$} \vspace{2mm} \\
$^1$Peking University  \quad  $^2$Nanjing University  \\
$^3$Microsoft Research Asia
}

\begin{document}
\maketitle

\blfootnote{$^*$: Equal contribution and work done during the internship at Microsoft Research Asia.}

\blfootnote{$^{\dagger}$: Project lead <xiaoyizhang@microsoft.com>. }

\begin{abstract}
GUI agents that interact with graphical interfaces on behalf of users represent a promising direction for practical AI assistants. However, training such agents is hindered by the scarcity of suitable environments. We present \textsc{InfiniteWeb}, a system that automatically generates functional web environments at scale for GUI agent training. While LLMs perform well on generating a single webpage, building a realistic and functional website with many interconnected pages faces challenges. We address these challenges through unified specification, task-centric test-driven development, and a combination of website seed with reference design image to ensure diversity. Our system also generates verifiable task evaluators enabling dense reward signals for reinforcement learning. Experiments show that \textsc{InfiniteWeb} surpasses commercial coding agents at realistic website construction, and GUI agents trained on our generated environments achieve significant performance improvements on OSWorld, Online-Mind2Web, and MobileWorld, demonstrating the effectiveness of the proposed system. Our code and generated websites are publicly available at \url{https://github.com/microsoft/FIVE-UI-Evol}.
\end{abstract}

%=============================================================================
\section{Introduction}
%=============================================================================

\begin{figure}[t]
\centering
\includegraphics[width=\columnwidth]{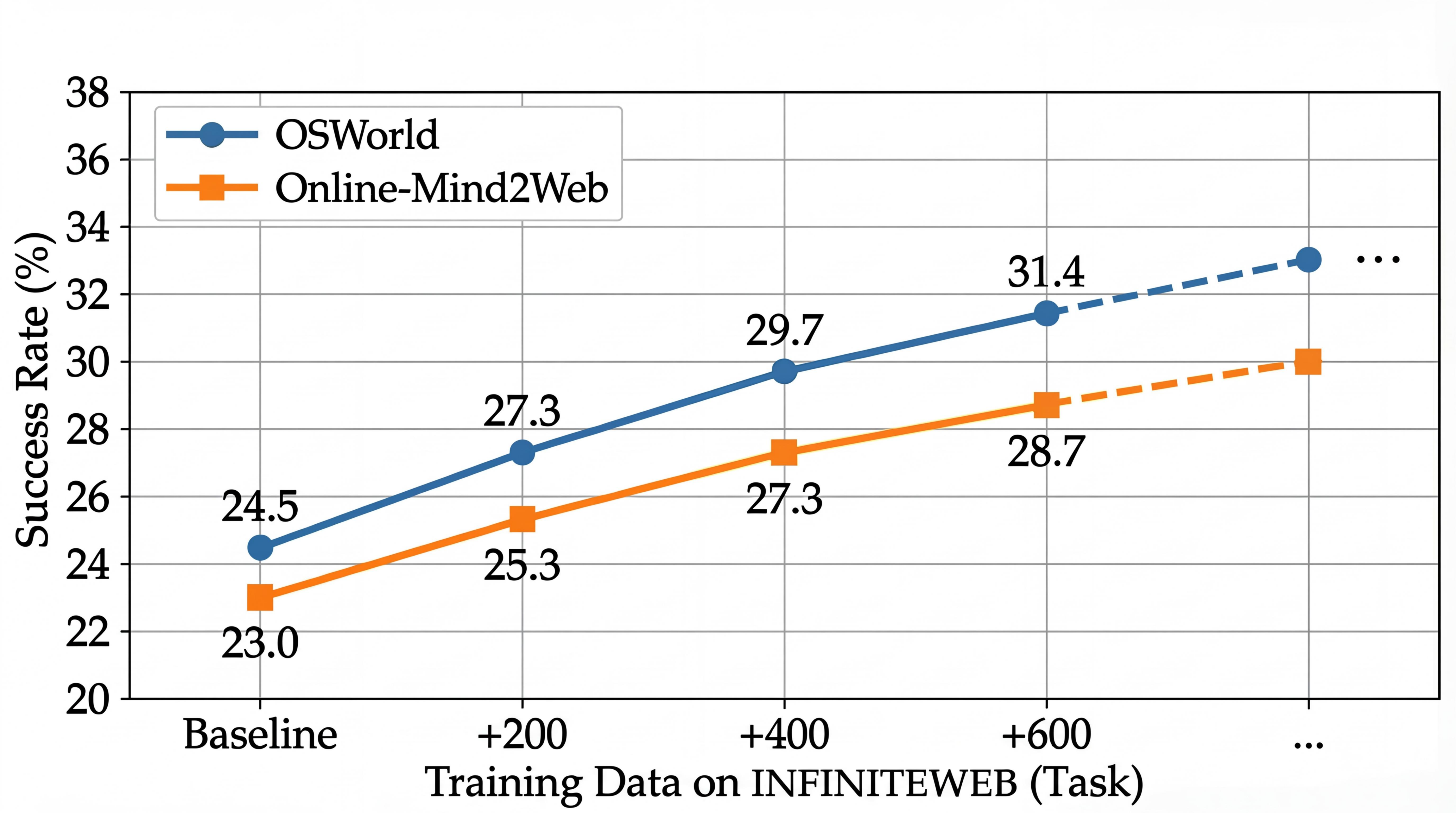}
\caption{GUI agent performance improves with more training data generated by \textsc{InfiniteWeb}. Dashed lines indicate potential for further scaling.}
\label{fig:scaling}
\end{figure}

\begin{figure*}[t]
\centering
\includegraphics[width=\textwidth]{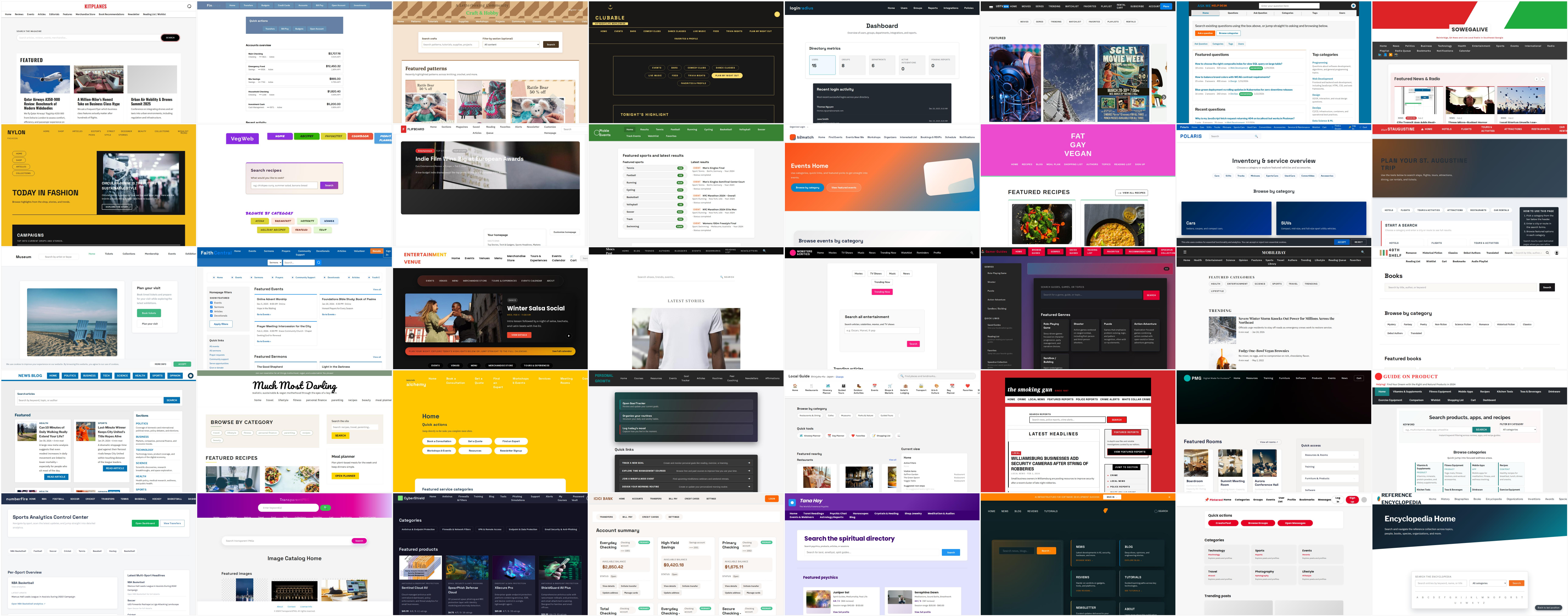}
\caption{A showcase of diverse websites generated by \textsc{InfiniteWeb}, spanning various categories including e-commerce, booking platforms, social media, and more.}
\label{fig:showcase}
\end{figure*}

GUI agents, autonomous systems that interact with graphical user interfaces to complete tasks on behalf of users, have emerged as a promising direction for building practical AI assistants \cite{xie2024osworld,zhou2024webarena}. Recent advances \cite{hong2024cogagent,qin2025uitars} have demonstrated that vision-language models can be end-to-end trained with reinforcement learning algorithms as GUI agents to understand screenshots, reason about UI elements, and execute human-like actions to automate tasks in digital world. However, training such agents remains challenging due to the scarcity of suitable  environments.

Existing GUI agent benchmarks, such as MiniWoB++ \cite{liu2018miniwob}, WebArena \cite{zhou2024webarena}, and OSWorld \cite{xie2024osworld}, provide valuable testbeds but suffer from fundamental limitations in \textbf{scale} and \textbf{diversity} as training environments. These benchmarks are manually constructed, requiring significant human effort to design websites or download applications, define tasks, and create evaluation criteria. As a result, they contain only tens to hundreds of applications, insufficient for training agents that can generalize across the vast diversity of real-world websites. Although recent work~\cite{sun2025genesis,xu2024agenttrek,xie2025agentsynth} proposes synthesizing tasks or trajectories, these approaches still operate within the same benchmark environments, limiting model training on a small set of specific applications.

A natural question arises: \textit{Can we automatically generate environments for GUI agent training?} While large language models (LLMs) have shown remarkable code generation capabilities~\cite{chen2022codet,si2024design2code, jimenez2023swe}, especially for web frontend~\cite{genui}, directly applying them to generate complete, functional websites faces three critical challenges.

Generating such environments presents three intertwined challenges. First, \textbf{consistency}: While LLMs perform well on generating a single webpage, a realistic website comprises multiple interconnected pages sharing data, visual styles, and backend interfaces. LLMs generating pages independently often produce incompatible implementations, different backend interface signatures, conflicting data formats, or inconsistent state management, breaking the cross-page interactions essential for realistic websites. Second, \textbf{correctness}: website functionalities require multiple coordinated steps, but LLM-generated code frequently contains functional bugs that compound over long-horizon tasks, causing incorrect reward signals that can destabilize reinforcement learning. Third, \textbf{diversity}: LLMs tend to produce repetitive task patterns and homogeneous visual styles, risking agent overfitting to specific interaction patterns rather than learning generalizable skills.

In this paper, we present \textsc{InfiniteWeb}, an agentic system that automatically generates functional web environments at scale for GUI agent training, addressing aforementioned challenges.

For \textbf{consistency}, we propose \textbf{Unified Specification}: rather than generating pages independently, we first derive a complete set of data models and interfaces from user tasks, then generate all pages according to this shared specification, ensuring the realistic cross-page interactions.
To ensure \textbf{correctness}, inspired by the classic software engineering practice \cite{williams2003tdd}, we introduce the \textbf{task-centric test-driven development (TCTDD)} approach, where test cases are firstly derived from task specifications and then code  is iteratively refined until all task-relevant tests pass. 
For \textbf{diversity}, our system addresses it from both functional and visual dimensions: functionally, by taking a \textit{website seed} (a brief description) and generating tasks specifically designed to match that seed. Visually, by providing reference design images, we use vision-language models to extract characteristics and generate websites that match the target style. It enables the leverage the millions of visually distinct websites available in resources like Common Crawl~\cite{commoncrawl} as an abundant source of diverse designs.

\begin{figure*}[t]
  \centering
  \includegraphics[width=\textwidth]{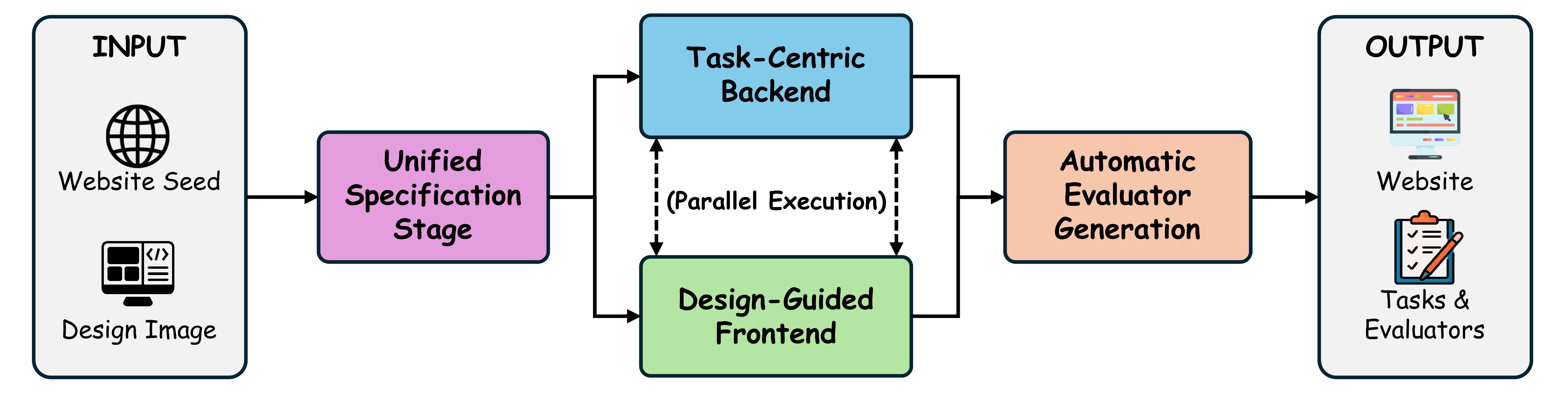}
  \caption{Overview of \textsc{InfiniteWeb}. Given a website seed and design image, our system produces a functional website with tasks and evaluators through four stages: the \textbf{Unified Specification Stage} generates tasks and derives data models and interfaces; the \textbf{Task-Centric Backend} and \textbf{Design-Guided Frontend} execute in parallel; and the \textbf{Evaluator Generation} creates task-specific evaluators for dense reward signals.}
  \label{fig:overview}
\end{figure*}

Furthermore, to support RL-based training, our system is designed to generate \textbf{verifiable task evaluators} along with the website and tasks, which tracks key task-related variables during agent running, enabling dense reward signals for reinforcement learning. We conduct systematical analysis on our system from two aspects: generated website quality and the effect to training GUI agent as simulated environment. The results demonstrate the superiority of our system as an environment synthesis system.

We summarize our contributions as follows and we will release the artifacts of this work to further contribute to the research community:
\begin{itemize}[leftmargin=*,itemsep=0mm]
  \item We propose \textsc{InfiniteWeb}, the first system specifically designed for generating functional web environments with verifiable evaluators for GUI agent training at scale (Figure~\ref{fig:showcase}).
  \item Experiments demonstrate that our system surpasses advanced coding agents in building realistic web environments on WebGen-Bench, achieving superior performance in both visual and functional quality.
  \item Training on our generated environments significantly improves GUI agent performance: +6.9\% on OSWorld, +5.7\% on Online-Mind2Web, and +3.9\% on MobileWorld, demonstrating the realism and quality of simulated environments produced by our system and their cross-platform transferability.
\end{itemize}

%=============================================================================
\section{Related Work}
%=============================================================================

\paragraph{GUI Agent Benchmarks.}
While there are benchmarks evaluating separate ability of GUI Agents like UI element grounding~\cite{li2025screenspot,liu2025ui} or UI understanding~\cite{wang2025mmbench},
end-to-end evaluating GUI agents requires interactive environments. 
Early work such as MiniWoB++ \cite{liu2018miniwob} introduced simplified web interaction tasks, demonstrating the potential of reinforcement learning for web automation. Subsequent benchmarks have increased realism and complexity: WebArena \cite{zhou2024webarena} provides self-hosted websites for autonomous agent evaluation, OSWorld \cite{xie2024osworld} extends to full desktop environments across multiple operating systems, and Mind2Web \cite{deng2023mind2web} offers large-scale web task annotations. However, these benchmarks share a fundamental limitation: they are manually constructed, requiring significant human effort to design environments, define tasks, and create evaluators. This limits their scale and diversity, potentially leading to agent overfitting. Our work addresses this bottleneck by automatically generating functional web environments at scale.

\paragraph{LLM-based Code and Website Generation.}
Large language models have shown remarkable code generation capabilities, from solving competitive programming problems \cite{li2022alphacode} to generating complete applications. Recent work has explored UI-to-code generation: Design2Code \cite{si2024design2code} benchmarks the conversion of visual designs to front-end code, while WebGen-Bench \cite{lu2025webgenbench} evaluates end-to-end website generation from natural language descriptions. However, a key challenge remains: LLM-generated code frequently contains bugs. CodeT \cite{chen2022codet} addresses this by generating tests alongside code to filter incorrect solutions. Our approach builds on this insight but differs in a crucial way: rather than attempting to verify all generated code, we focus on \textit{task-centric correctness}, ensuring only the functionality required for specific user tasks is bug-free, making the verification problem tractable.

\paragraph{Synthetic Environment and Data Generation.}
Procedural generation has proven valuable for training robust agents. \citet{cobbe2020procgen} demonstrated that procedurally generated game levels significantly improve reinforcement learning generalization. In the GUI agent domain, recent work has explored synthetic data generation: WebSailor-V2 \cite{li2025websailor} uses synthetic trajectories and scalable RL to train web agents, UI-Evol \cite{zhang2025uievol} automatically evolves knowledge for computer use agents, while AgentSynth \cite{xie2025agentsynth} synthesizes long-horizon desktop tasks from atomic subtasks. These approaches focus on generating \textit{training data} (action trajectories) within existing environments. In contrast, our work generates complete, functional \textit{environments} themselves, including websites, tasks, and automatic evaluators, addressing the environment scalability problem at its source.

\begin{figure}[t]
  \centering
  \includegraphics[width=\columnwidth]{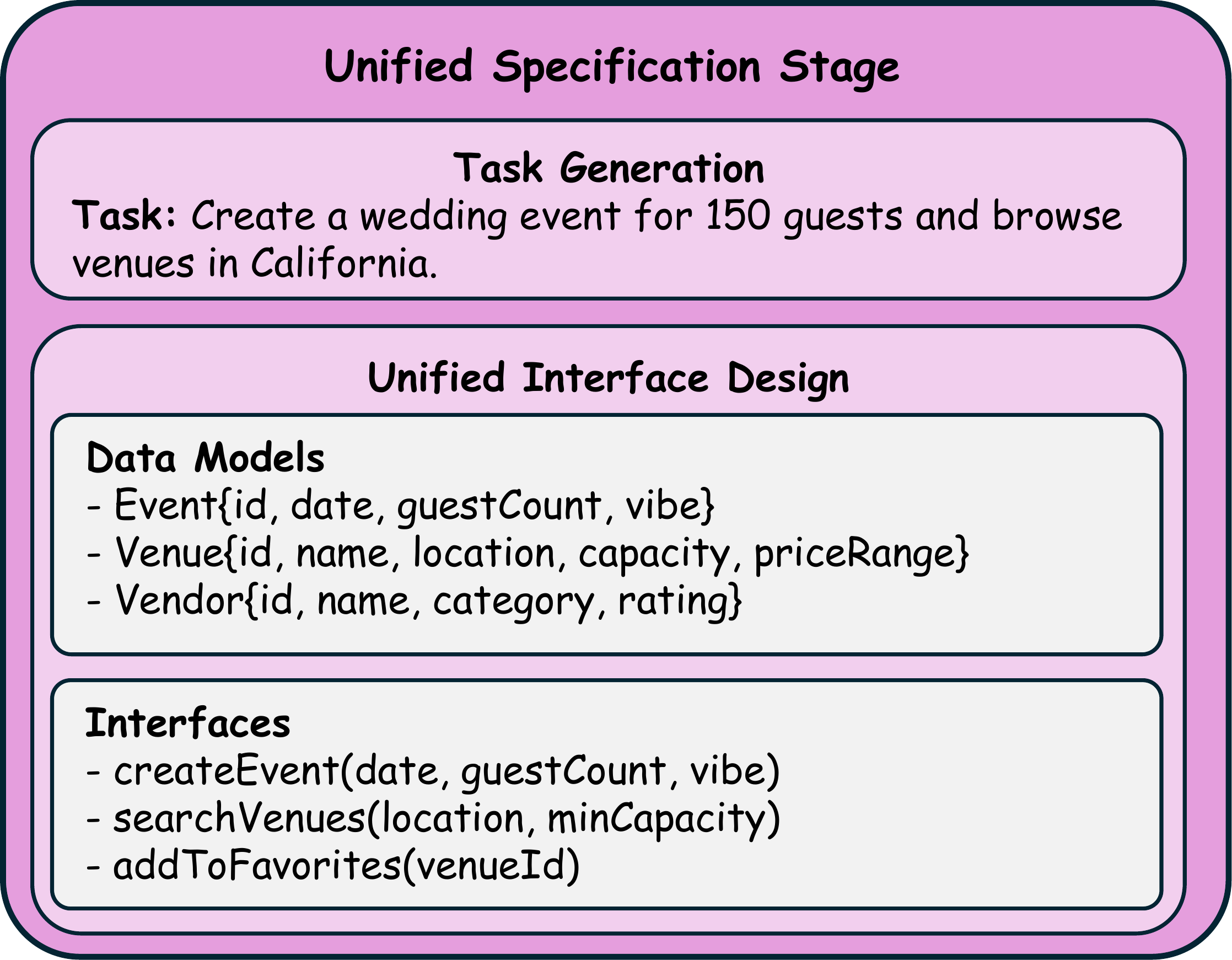}
  \caption{Unified Specification Stage. Given a website seed and design image, this stage generates realistic tasks, then derives shared interface design consisting of data models and programming interfaces across pages.}
  \label{fig:specification}
\end{figure}

%=============================================================================
\section{Method}
%=============================================================================

\subsection{Overview}

Figure~\ref{fig:overview} illustrates our system pipeline. Our system takes a website seed (e.g., ``online bookstore website'') and a design image as input, and outputs a fully functional website along with tasks that can be done in the website and corresponding automatic evaluators. Both website seeds and design images are extracted from Common Crawl to provide diverse visual and functional references (details in Appendix~\ref{sec:appendix-design-image}).

Our pipeline consists of four main stages, with the backend and frontend executing in parallel. First, the \textbf{Unified Specification Stage} generates tasks and derives unified data models and interfaces, ensuring \textbf{consistency} and \textbf{functional diversity}. Second, the \textbf{Task-Centric Backend} uses TCTDD to validate business logic, ensuring \textbf{correctness} of task-relevant functionality. Third, the \textbf{Design-Guided Frontend} extracts visual features from design images to guide page generation, ensuring \textbf{visual diversity}. Fourth, \textbf{Evaluator Generation} produces task-specific evaluators with \textbf{dense reward signals} for reinforcement learning.

%-----------------------------------------------------------------------------
\subsection{Unified Specification Stage}
\label{sec:prepare}
%-----------------------------------------------------------------------------

This stage addresses the \textbf{consistency} challenge while enabling \textbf{functional diversity}. A functional website typically consists of multiple pages that share data and interfaces. When generating pages independently, LLMs often produce inconsistent implementations. Our key insight is that \textit{everything should be derived from tasks}: by first generating tasks specific to the website seed, then deriving unified data models and interfaces from them, we ensure all pages share identical specifications while tasks naturally vary across different website seeds.
\begin{figure}[t]
  \centering
  \includegraphics[width=\columnwidth]{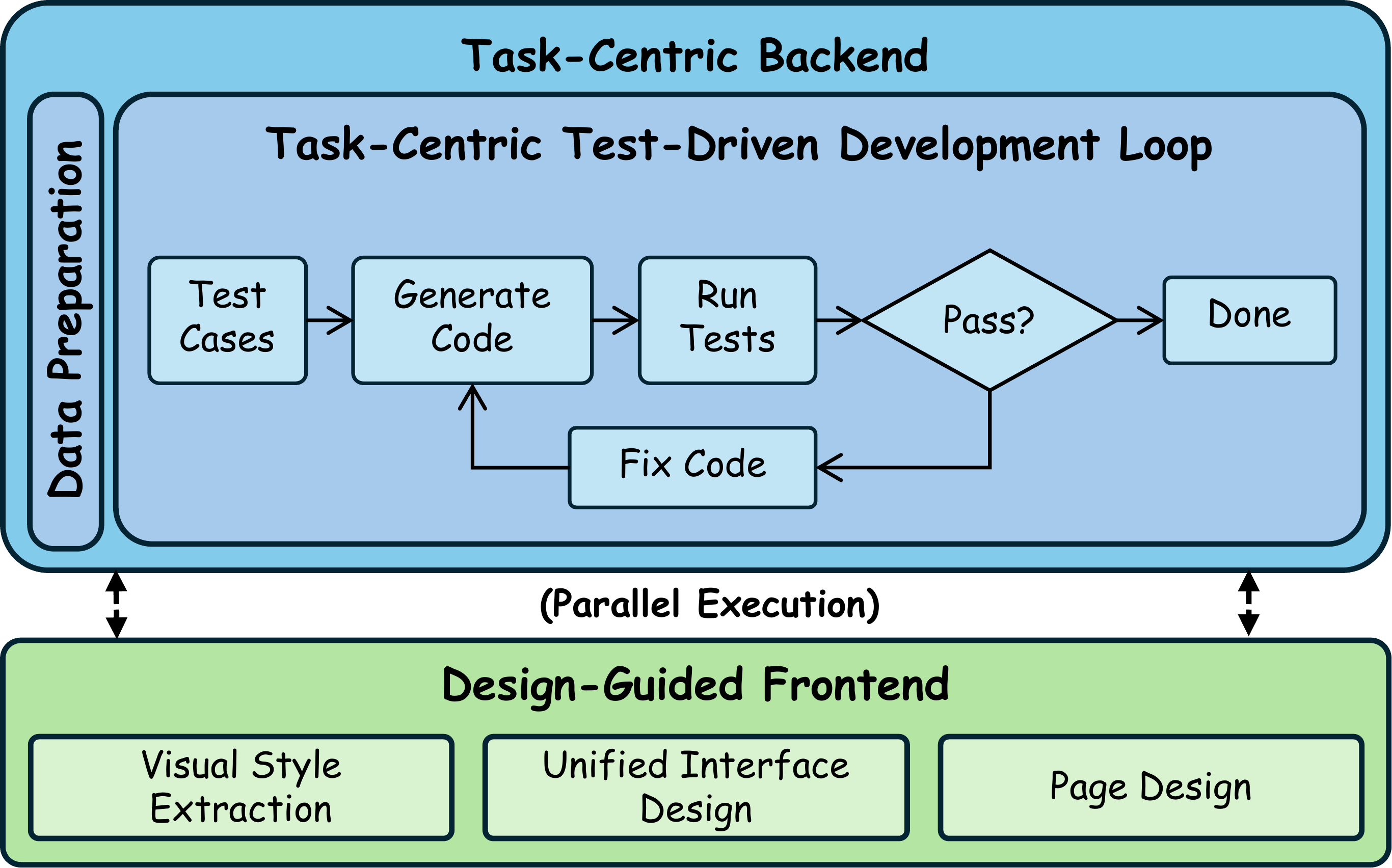}
  \caption{Task-Centric Backend and Design-Guided Frontend in parallel. The backend uses TCTDD to iteratively generate and validate business logic. The frontend extracts visual styles and generates pages.}
  \label{fig:parallel}
\end{figure}

\paragraph{Task Generation.}
Given a website seed (e.g., ``online bookstore''), we prompt an LLM to generate realistic user tasks specific to that website seed. This ensures \textbf{functional diversity}: a booking website generates reservation-related tasks (e.g., ``book a hotel room for next weekend''), while an e-commerce site generates shopping tasks (e.g., ``find and purchase a laptop under \$500''). Each task represents a complete user goal that varies in complexity and covers different aspects of the website's functionality.

\paragraph{Unified Interface Design.}
From the generated tasks, we derive three unified specifications that all pages share. First, we extract \textit{data models}: if tasks involve searching products, viewing details, and making purchases, we derive entities such as \texttt{Product}, \texttt{Cart}, and \texttt{Order} with their attributes and relationships. Second, we perform \textit{preliminary architecture planning} to identify all pages required (e.g., homepage, search results, product details, cart, checkout) and their primary functions. Third, we derive a unified set of \textit{programming interfaces}: each task step implies one or more interface calls, and crucially, these interface specifications are shared across all pages, ensuring identical parameters and data formats.

The interfaces are designed to be \textit{user-facing}: the system automatically classifies parameters into system-managed (e.g., \texttt{userId}, \texttt{sessionId}, managed internally) and user-provided (e.g., \texttt{productId}, \texttt{quantity}). For example, the original interface \texttt{addToCart(userId, sessionId, productId, quantity)} is wrapped as \texttt{addToCart(productId, quantity)}, with system parameters automatically retrieved from localStorage. This unified interface design ensures that all pages use identical API signatures and data formats, enabling seamless cross-page interactions.

With the unified specification stage complete (tasks, data models, interfaces), we now turn to generating business logic and frontend pages in two parallel pipelines.

%-----------------------------------------------------------------------------
\subsection{Task-Centric Backend}
\label{sec:backend}
%-----------------------------------------------------------------------------

This stage addresses the \textbf{correctness} challenge. LLM-generated code frequently contains bugs, making naively synthesized environments unsuitable for agent learning. Our key insight is to adopt \textit{task-centric correctness} as the correctness criterion. Since agents interact only with a narrow, task-induced subspace determined by task specifications and their policies, correctness outside this subspace does not contribute to the learning signal or policy optimization. Rather than enforcing full functional correctness over the entire website, we focus on ensuring that only the functionalities required for the target tasks are correct. This alignment allows correctness verification and refinement to be focused on task-relevant execution paths, which we operationalize through TCTDD.

\paragraph{Data Preparation.}
We generate concrete data instances that populate the website, ensuring consistency with both data models and tasks. For example, if a task requires finding products under \$50, we ensure the generated product catalog contains such items. Placeholder resources (e.g., image URLs) are replaced with real, context-relevant content via external APIs.

\paragraph{Task-Centric Test-Driven Development.}
We adopt the TCTDD approach to ensure correctness of task-relevant functionality. TCTDD works as follows: based on task specifications and generated data, test cases and implementation code are generated in parallel; then tests are run and iteratively fixed until all pass.

Test cases and implementation code use the same pre-generated data, ensuring consistency. For example, if the generated data contains products priced at \$29.99 and \$45.00, the test will verify that exactly these products are returned for an ``under \$50'' query. When tests fail, we provide the LLM with the failing test case, expected vs. actual output, and relevant code segment. The LLM generates a fix and re-tests, continuing until all tests pass or a maximum iteration limit is reached.

%-----------------------------------------------------------------------------
\subsection{Design-Guided Frontend}
\label{sec:frontend}
%-----------------------------------------------------------------------------

This stage is designed to address the \textbf{visual diversity} challenge. LLMs tend to generate websites with similar visual styles. Our key insight is to draw \textit{design images} as references from abundant and visually diverse website screenshot in the real world. Given a reference design image, we extract visual characteristics and generate pages that match the guidance.

\paragraph{Visual Style Extraction.}
We employ a vision-language model to decode visual attributes from the design image, establishing a global style constraint. Specifically, we extract the \textbf{color system} (\textit{e.g.}, primary and neutral palettes), \textbf{typography hierarchy} including font families and weights, \textbf{spacing rules}, and \textbf{component styling} such as button patterns. These extracted structural style specification  serve as a consistent visual specification for all subsequent generation steps.

\paragraph{Page Design.}
Building on the unified specification, we conduct detailed architectural design for each page in parallel. This step defines specific \textbf{functional requirements} including content blocks and interaction flows, determines \textbf{routing logic} via URL parameters, and establishes \textbf{responsive layouts} defined by grid systems and breakpoints.

\paragraph{Page Realization.}
We first generate a unified page framework containing the shared header, footer, and CSS variables for the website, based on the extracted visual features, ensuring consistent styling across all pages. Then, for each page, we generate the HTML structure, CSS styles, and a JavaScript UI layer that connects elements to the backend SDK using a data-attribute–driven pattern (e.g., \texttt{data-populate}, \texttt{data-action}). Finally, we inject an initialization script into the homepage that writes the generated data to \texttt{localStorage}, which is the browser’s built-in persistent key–value storage. It enables data persistence across pages without a backend server.

\begin{table*}[t]
\centering
% \small
% \setlength{\tabcolsep}{7pt}
\resizebox{\textwidth}{!}{
\begin{tabular}{lcccccccc}
\toprule
\multirow{2}{*}{\textbf{Method}} & \multicolumn{3}{c}{\textbf{Instruction Categories}} & \multicolumn{3}{c}{\textbf{Test Case Categories}} & \multirow{2}{*}{\textbf{Overall}} \\
\cmidrule(lr){2-4} \cmidrule(lr){5-7}
 & Content Pres. & User Inter. & Data Mgmt. & Functional & Data Display & Design Valid. & \\
\midrule
Bolt.diy & 83.2 & 59.9 & 62.8 & 58.4 & 84.1 & 41.7 & 67.0 \\
Claude-Code & 87.9 & 70.1 & 67.6 & 67.3 & 87.5 & 61.1 & 74.3 \\
Codex & 89.8 & 79.2 & 75.6 & 72.8 & \textbf{96.2} & 76.4 & 81.2 \\
\textbf{Ours} & \textbf{91.5} & \textbf{83.8} & \textbf{82.7} & \textbf{80.9} & 94.1 & \textbf{82.8} & \textbf{85.6} \\
\bottomrule
\end{tabular}
}
\caption{Category-wise evaluation results on WebGen-Bench (\%). Instruction Categories classify the website functionality type. Test Case Categories classify the evaluation type. Results are averaged over three runs.}
\label{tab:category-results}
\end{table*}

\begin{table*}[t]
\centering
\small
\setlength{\tabcolsep}{3pt}
\begin{tabular}{lccccccccccc}
\toprule
\textbf{Method} & \textbf{Chrome} & \textbf{GIMP} & \textbf{Calc} & \textbf{Impress} & \textbf{Writer} & \textbf{Multi} & \textbf{OS} & \textbf{Thunder.} & \textbf{VLC} & \textbf{VSCode} & \textbf{Overall} \\
\midrule
Computer-use-preview & 36.9 & 34.6 & 10.6 & 25.4 & 30.4 & 10.8 & 45.8 & 46.7 & 29.4 & 47.8 & 26.0 \\
Claude-3.7-Sonnet & 41.2 & 34.6 & 8.5 & 29.7 & 39.1 & 10.8 & 50.0 & 33.3 & 35.3 & 43.5 & 27.1 \\
Doubao-1.5-thinking-0428 & 44.4 & 46.2 & 13.0 & 31.8 & 39.1 & 4.8 & 30.4 & 66.7 & 23.5 & 56.5 & 28.1 \\
Claude-4-Sonnet & 36.9 & 46.2 & 17.0 & 36.2 & 43.5 & 9.7 & 37.5 & 66.7 & 38.5 & 60.9 & 31.2 \\
OpenCUA-32B & 40.5 & 55.1 & 13.5 & 30.7 & 39.1 & 9.7 & 52.2 & 44.4 & 25.0 & 52.8 & 29.7$\pm$1.1 \\
\midrule
UI-TARS-1.5-7B & 22.9 & 51.9 & 11.7 & 29.7 & 39.1 & 3.8 & 34.8 & 26.7 & 34.2 & 63.0 & 24.5$\pm$1.2 \\
\ \ + 200 tasks & 34.8 & 61.5 & 10.0 & 27.7 & 34.8 & 8.0 & 41.7 & {40.0} & 25.5 & 55.1 & 27.3$\pm$0.9 \\
\ \ + 400 tasks & 35.5 & {69.2} & 10.6 & {29.8} & 37.7 & 9.0 & {48.6} & 35.5 & 33.3 & 62.3 & 29.7$\pm$1.1 \\
\ \ + 600 tasks & {36.9} & {69.2} & {12.8} & {29.8} & {47.8} & {9.7} & 45.8 & {40.0} & {35.3} & {65.2} & \textbf{31.4$\pm$1.0} \\
\bottomrule
\end{tabular}
\caption{Results on OSWorld under 15 maximum steps by domain (\%). The lower section shows UI-TARS-1.5-7B trained with tasks from \textsc{InfiniteWeb}-generated websites. Calc/Impress/Writer refer to LibreOffice applications. Multi = Multi-Apps, Thunder. = Thunderbird. Standard deviation computed over three runs.}
\label{tab:osworld}
\end{table*}

%-----------------------------------------------------------------------------
\subsection{Automatic Evaluator Generation}
\label{sec:evaluator}
%-----------------------------------------------------------------------------

A critical requirement for GUI agent training is automatically evaluating whether a task has been successfully completed. Our system automatically generates task-specific evaluators by leveraging existing state variables and code instrumentation.

The evaluators leverage two types of variables: \textit{existing variables} representing state naturally stored by the application (e.g., cart contents, user preferences), and \textit{instrumentation variables} that are explicitly added checkpoints tracking task-specific progress. For instrumentation variables, we identify the key steps required for each task's completion and record progress in \texttt{localStorage} when the corresponding functions execute. For example, a ``search and purchase'' task might track: search query submitted, product viewed, item added to cart, and checkout completed.

Based on these variables, we generate a JavaScript evaluator function that checks variables to determine task completion, capable of assessing partial completion rather than only binary success/failure. This enables \textit{dense reward signals} for reinforcement learning: agents receive partial credit based on completed steps, facilitating more effective learning for complex multi-step tasks.

%=============================================================================
\section{Experiments}
\label{sec:experiments}
%=============================================================================

We evaluate \textsc{InfiniteWeb} on three dimensions: (1) functional correctness of generated websites, (2) visual quality, and (3) effectiveness for GUI agent training.

%-----------------------------------------------------------------------------
\subsection{Experimental Setup}
%-----------------------------------------------------------------------------

We evaluate on four benchmarks and assess visual quality through pairwise comparisons. For fair comparison, all website generation methods use GPT-5 as the backbone LLM with reasoning effort set to ``high''. Implementation details including generation hyperparameters and agent training configuration are provided in Appendix~\ref{sec:appendix-impl}.  We also provide manual evaluation detailed in Appendix~\ref{sec:appendix-task-quality}.

% \paragraph{WebGen-Bench.}
% WebGen-Bench \cite{lu2025webgenbench} evaluates functional correctness of LLM-generated websites through agent-based task execution on 101 websites. Each website is generated with a set of predefined user tasks that it must support. For evaluation, an LLM agent is presented with a user task (e.g., ``search for products under \$50 and add the cheapest one to cart'') and attempts to complete it by interacting with the generated website. The agent can perform up to 15 sequential actions (clicks, form inputs, navigation) before the evaluation terminates. Each task outcome is classified as: \textit{Passed} if the task is fully completed with correct results, \textit{Partial} if the agent makes progress but does not complete the task entirely, or \textit{Failed} if the task cannot be accomplished due to missing functionality or errors. We report three metrics: Passed rate, Partial rate, and the Overall score. Since the original WebGen-Bench does not include design images, we match each test website with a design image extracted from Common Crawl based on website category. This design image is provided to all methods as input to enable fair comparison.

\paragraph{WebGen-Bench.}
WebGen-Bench \cite{lu2025webgenbench} evaluates functional correctness of LLM-generated websites through agent-based task execution on 101 websites. Each website is generated with a set of predefined tasks that it must support. For evaluation, an LLM agent is presented with a user task and attempts to complete it by interacting with the generated website. 
% The agent can perform up to 15 sequential actions (clicks, form inputs, navigation) before the evaluation terminates. 
Each task outcome is classified as: \textit{Passed} if the task is fully completed with correct results, \textit{Partial} if the agent makes progress but does not complete the task entirely, or \textit{Failed} if the task cannot be accomplished due to missing functionality or errors. We report three metrics: Passed rate, Partial rate, and the Overall score. Since the original WebGen-Bench does not include design images, we match each test website with a design image extracted from Common Crawl based on website category. This design image is provided to all methods as input to enable fair comparison.
\begin{figure}[t]
\centering
\includegraphics[width=\columnwidth]{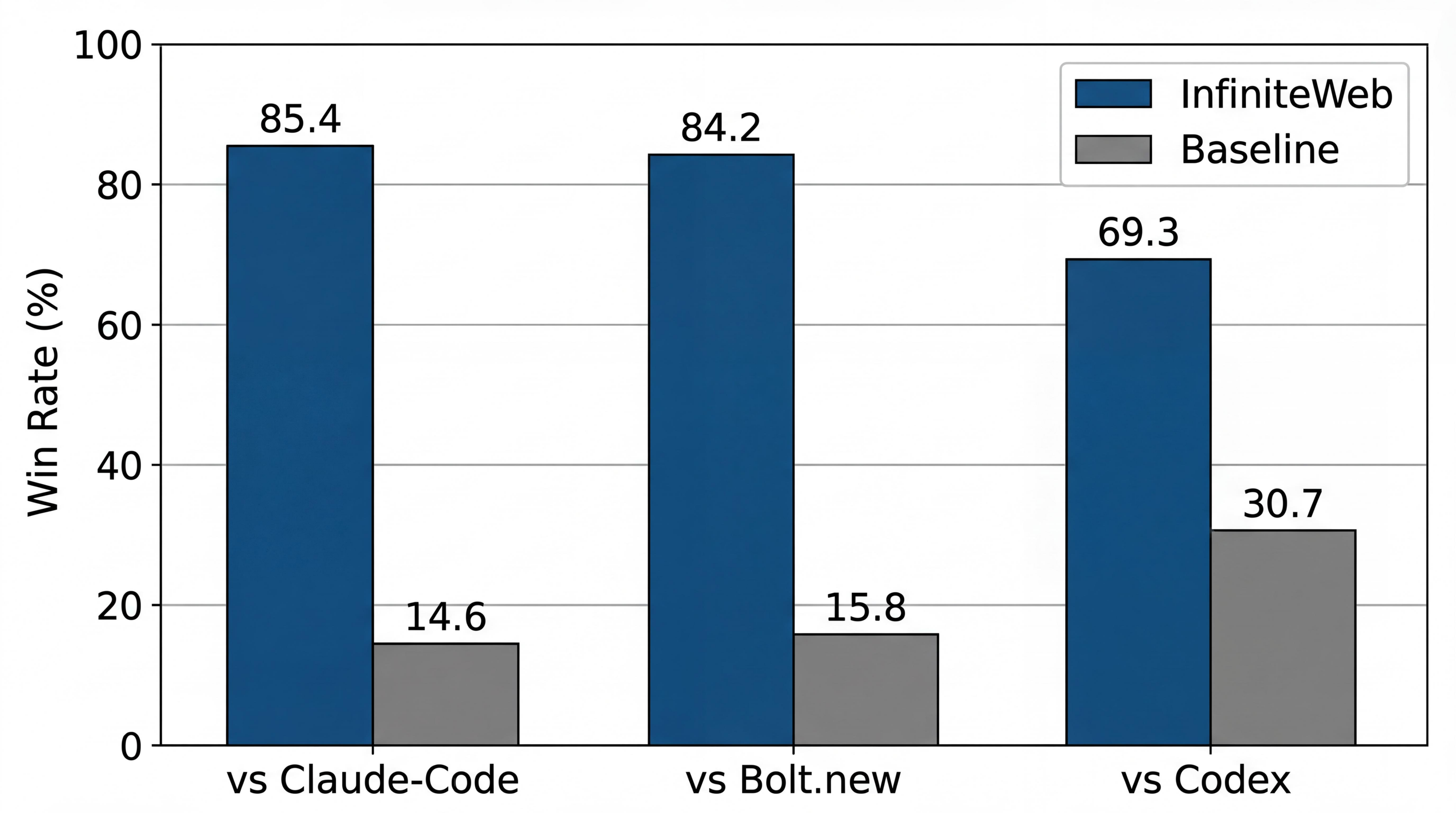}
\caption{LLM-as-Judge visual quality evaluation. Each pair shows win rates for ours (left) vs baseline (right).}
\label{fig:visual}
\end{figure}

\paragraph{LLM-as-Judge Visual Quality.}
We assess visual quality through LLM-as-Judge pairwise comparisons \cite{zheng2023judging} on 200 generated websites. For each website, we capture a full-page screenshot and present it alongside the reference design image to GPT-5. The model is prompted to evaluate which implementation better matches the target design across five dimensions: (1) visual layout similarity, (2) color scheme matching, (3) typography and spacing, (4) component arrangement and structure, and (5) overall aesthetic consistency. The model outputs one of three judgments: our method wins, the baseline wins, or tie. We report win rates for each pairwise comparison, where higher percentages indicate stronger visual fidelity to the design reference.

\paragraph{Online-Mind2Web.}
Online-Mind2Web \cite{xue2025onlinemind2web} extends the original Mind2Web benchmark \cite{deng2023mind2web} to evaluate web agents on live websites, testing their ability to complete realistic tasks on real-world web pages. Unlike static benchmarks with cached HTML snapshots, Online-Mind2Web requires agents to interact with actual deployed websites, introducing challenges such as dynamic content loading, varying page layouts, and real network latency. We use this benchmark to measure \textbf{in-domain generalization}: whether training on our synthetic websites improves performance on real-world web interactions that the agent has never seen during training.

\paragraph{OSWorld.}
OSWorld \cite{xie2024osworld} is a benchmark for evaluating GUI agents on real desktop applications across diverse domains including web browsers, office suites (Calc, Impress, Writer), media players (VLC), code editors (VS Code), and email clients (Thunderbird). We use this benchmark to measure \textbf{out-of-domain transfer}: whether training on synthetic web environments transfers to real desktop application tasks. Specifically, we adopt \textbf{OSWorld-Verified}\cite{osworld_verified}, a refined version with improved task quality and evaluation robustness.

\paragraph{MobileWorld.}
MobileWorld \cite{kong2025mobileworld} benchmarks autonomous mobile agents in interactive environments. We use this benchmark to measure \textbf{mobile transfer}: whether training on our generated web environments, which use responsive layout design, transfers to mobile GUI tasks without additional fine-tuning.

%-----------------------------------------------------------------------------
\subsection{Website Functional Correctness}
\label{sec:functional}
%-----------------------------------------------------------------------------

We compare against three representative approaches for AI-powered website generation: \textbf{Codex} (v0.46.0) \cite{chen2021codex}, OpenAI's coding assistant agent; \textbf{Claude-Code} (v2.0.0) \cite{anthropic2025claudecode}, Anthropic's coding assistant agent; and \textbf{Bolt.diy} (v0.0.7) \cite{boltdiy2024}, an open-source AI website builder from StackBlitz. All methods are given the same website seed and a homepage design image. The prompt template for baselines is provided in Appendix~\ref{sec:appendix-prompts}.

Table~\ref{tab:category-results} presents the functional correctness results on WebGen-Bench. Our method achieves the highest overall score of 85.6\%, significantly outperforming all baselines. We report performance across two classification schemes: \textit{Instruction Categories} (Content Presentation, User Interaction, Data Management) that classify the type of website functionality being tested, and \textit{Test Case Categories} (Functional Testing, Data Display Testing, Design Validation Testing) that classify the type of evaluation being performed. Our method achieves the best performance in Functional Testing (80.9\%) and Design Validation (82.8\%), demonstrating particularly strong advantages on the most challenging task categories. Detailed results with statistical significance tests are provided in Appendix~\ref{sec:appendix-results}.

\begin{figure*}[t]
\centering
\begin{minipage}{0.48\textwidth}
\centering
\includegraphics[width=\linewidth]{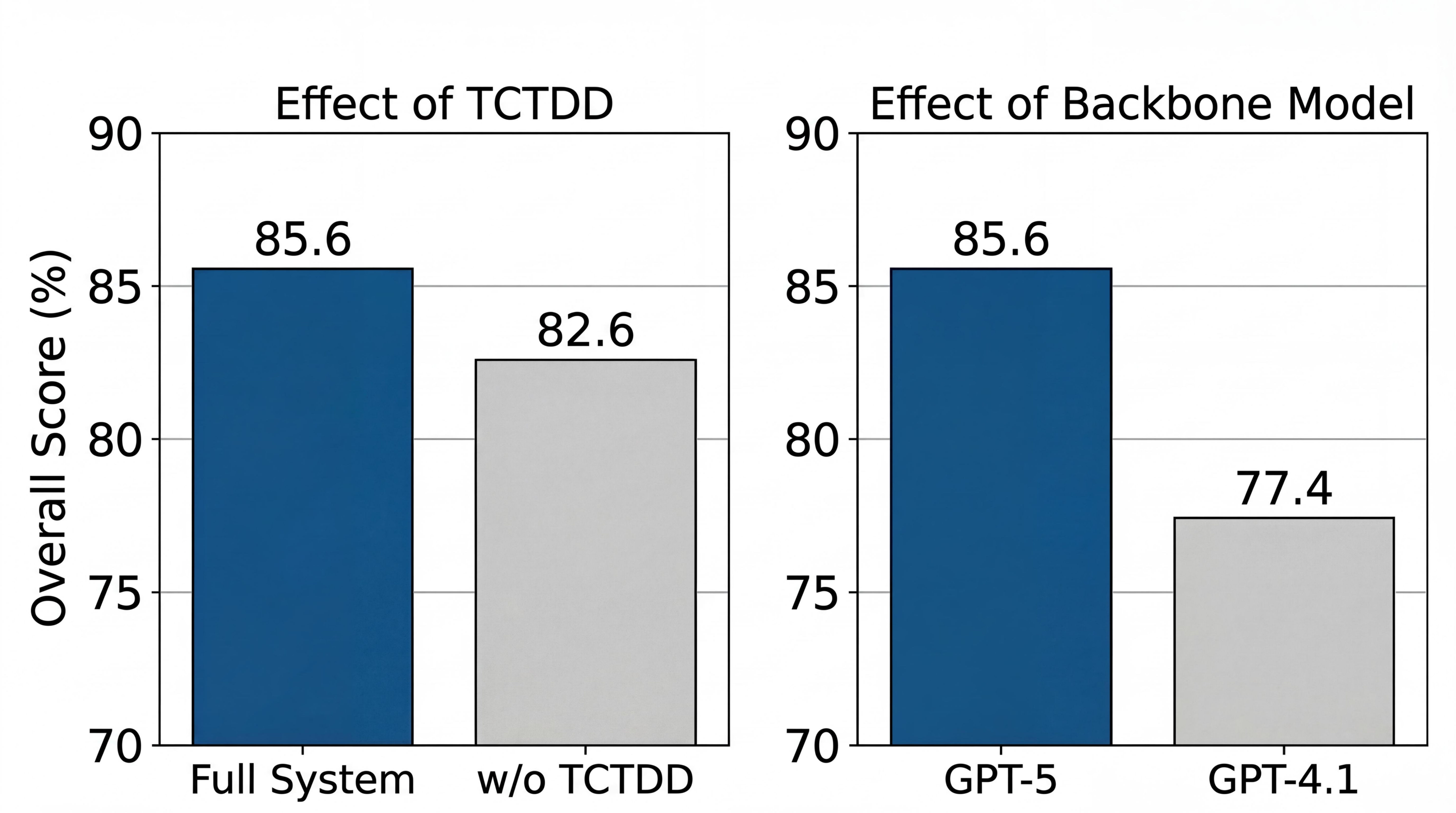}
\caption{Ablation study results on WebGen-Bench. Left: Effect of TCTDD validation loop. Right: Effect of backbone model.}
\label{fig:ablation}
\end{minipage}
\hfill
\begin{minipage}{0.48\textwidth}
\centering
\includegraphics[width=\linewidth]{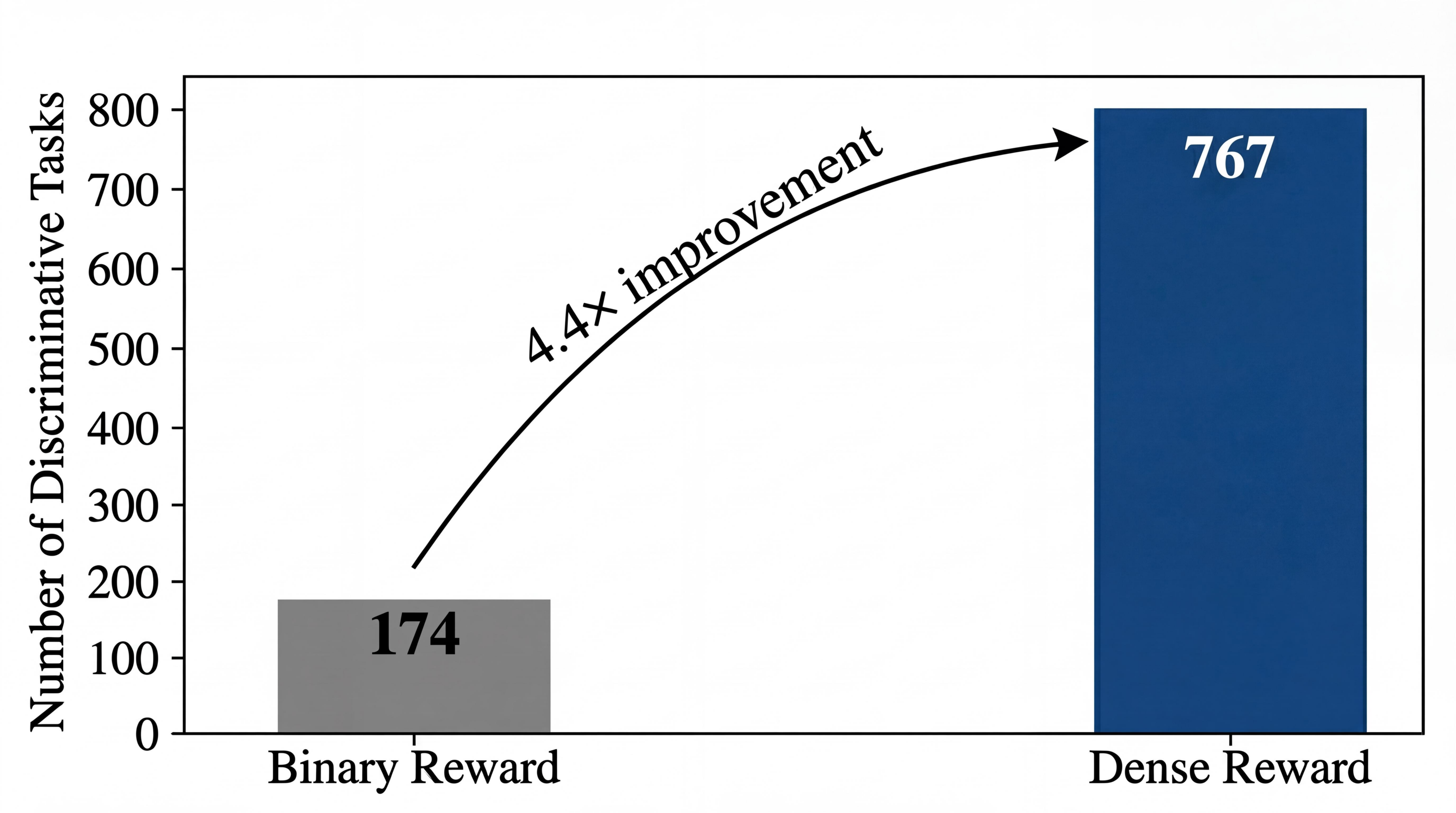}
\caption{Number of discriminative tasks for GRPO training. Dense reward enables learning from 4.4$\times$ more tasks by providing partial credit for intermediate steps.}
\label{fig:dense-reward}
\end{minipage}
\vspace{-4mm}
\end{figure*}

%-----------------------------------------------------------------------------
\subsection{LLM-as-Judge Visual Quality}
%-----------------------------------------------------------------------------

We compare the same websites generated in Section~\ref{sec:functional} (ours vs. Codex, Claude-Code, and Bolt.diy). Figure~\ref{fig:visual} shows the pairwise comparison results. Our method consistently outperforms all baselines (69--85\% win rate). Human evaluation confirms 91\% agreement with automated assessments (Appendix~\ref{sec:appendix-human}).

%-----------------------------------------------------------------------------
\subsection{Effectiveness for Agent Training}
%-----------------------------------------------------------------------------

The ultimate goal of \textsc{InfiniteWeb} is to provide training environments for GUI agents. We generate 600 tasks spanning diverse website categories (e-commerce, social media, booking platforms, etc.) and use them to train UI-TARS-1.5-7B \cite{qin2025uitars}. Training uses GRPO (Group Relative Policy Optimization) \cite{shao2024deepseekmath} with our dense reward signals from code instrumentation, enabling the agent to receive partial credit for intermediate progress rather than binary success/failure. We then evaluate on Online-Mind2Web (in-domain), OSWorld (out-of-domain), and MobileWorld (mobile transfer) benchmarks.

As shown in Figure~\ref{fig:scaling}, training on our generated environments leads to substantial improvements: +6.9\% on OSWorld (24.5\% $\rightarrow$ 31.4\%), +5.7\% on Online-Mind2Web (23.0\% $\rightarrow$ 28.7\%), and +3.9\% on MobileWorld (6.4\% $\rightarrow$ 10.3\%). Table~\ref{tab:osworld} shows the per-domain breakdown on OSWorld, where improvements are observed across most application categories. This suggests that skills acquired from training on web environments can transfer beyond web tasks to both desktop and mobile applications. Appendix~\ref{sec:appendix-case} provides case studies analyzing this transfer. We also compare against training on Claude-Code-generated websites using the same setup; results in Appendix~\ref{sec:appendix-claude-code} show that data quality, not quantity, drives the improvement.

The improvement scales with the amount of training data, suggesting that generating more diverse environments could yield further gains.

%-----------------------------------------------------------------------------
\subsection{Generated Environment Quality}
%-----------------------------------------------------------------------------

To evaluate the quality of our generated environments, we compare the success rate and average successful steps of two agents on InfiniteWeb and OSWorld: UI-TARS-1.5-7B and Agent S2 \cite{agashe2025agents2}, a multi-agent system using GPT-4.1 as planner and UI-TARS-72B for grounding. Table~\ref{tab:env-quality} shows the results.

\begin{table}[h]
\centering
\small
\begin{tabular}{@{}llcc@{}}
\toprule
\textbf{Env.} & \textbf{Agent} & \textbf{Score} & \textbf{Steps} \\
\midrule
\multirow{2}{*}{InfiniteWeb} & UI-TARS-1.5-7B & 7.4 & 10.3 \\
 & Agent S2 & 14.1 & 10.9 \\
\midrule
\multirow{2}{*}{OSWorld} & UI-TARS-1.5-7B & 24.5 & 9.0 \\
 & Agent S2 & 27.3 & 9.3 \\
\bottomrule
\end{tabular}
\caption{Agent performance on InfiniteWeb and OSWorld. Score is the average task completion rate (\%). Steps is the average steps for successful tasks.}
\label{tab:env-quality}
\vspace{-4mm}
\end{table}

\paragraph{Higher Difficulty.}
Compared to OSWorld, InfiniteWeb is markedly more challenging: agents achieve 2--3$\times$ lower scores, and successful tasks require longer trajectories, suggesting increased task complexity.

\paragraph{Better Discriminability.}
Performance on InfiniteWeb is more sensitive to agent capability, resulting in a 6.7 percentage point gap between Agent S2 and UI-TARS, compared to 2.8 on OSWorld.

%-----------------------------------------------------------------------------
\subsection{Ablation Studies}
%-----------------------------------------------------------------------------

Having established the effectiveness of our full system, we now examine the contribution of individual components through ablation studies. Figure~\ref{fig:ablation} shows the results.

\paragraph{Effect of TCTDD.}
Removing the TCTDD validation loop reduces the overall score by 5.0 points. This confirms that iterative test-driven refinement is crucial for achieving high functional correctness, even when using a strong backbone model. Notably, even without TCTDD, our method still achieves 80.6\%, comparable to Codex, showing that our base architecture is itself competitive.

\paragraph{Effect of Backbone Model.}
Replacing GPT-5 with GPT-4.1 reduces the score by 8.2 points (85.6 $\rightarrow$ 77.4). Even with GPT-4.1, our method still outperforms Claude-Code using GPT-5 (75.8\%), showing that our approach remains competitive even with a weaker backbone model.

\paragraph{Effect of Dense Reward.}
Our instrumentation system enables dense reward signals by tracking intermediate task steps. To evaluate its impact on reinforcement learning, we run UI-TARS-1.5-7B on 4,000 generated tasks with 4 trajectories per task and compare the number of discriminative tasks where GRPO can effectively learn, i.e., tasks where at least one trajectory in a group receives different scores. As shown in Figure~\ref{fig:dense-reward}, dense reward enables learning from 767 tasks compared to 174 with binary reward, a 4.4$\times$ increase. This demonstrates that dense reward substantially expands the effective training signal by providing partial credit for intermediate progress, thereby improving training data efficiency.

%-----------------------------------------------------------------------------
\subsection{Generation Efficiency}
%-----------------------------------------------------------------------------

We analyze the computational cost of website generation. On average, generating a single website consumes approximately 0.36M input tokens and 0.34M output tokens. Using GPT-5 batch processing pricing (\$0.625/M input, \$5.00/M output), this translates to approximately \$1.93 per website. The median generation time is approximately 20 minutes per website with our API configuration, though this is highly dependent on API response speed and rate limits. Since each website is generated independently, multiple websites can be generated in parallel to increase throughput.

%=============================================================================
\section{Conclusion}
%=============================================================================

We presented \textsc{InfiniteWeb}, a system that aims to generate functional web environments for GUI agent training, addressing consistency through unified interface design, correctness through task-centric test-driven development, and diversity through website seed variation and design image guidance. 
Our system surpasses commercial coding agents in this scenario and experiment results demonstrate its advantages for training GUI agents.
By releasing our system and generated datasets, we hope to support future research in building more capable and generalizable GUI agents.

%=============================================================================
\section*{Limitations}
%=============================================================================

Our work has several limitations that suggest directions for future research.

\paragraph{Single-Website Scope.} Our current tasks operate within individual websites. Cross-website tasks, such as comparing prices across multiple shopping sites or aggregating information from different sources, represent an important real-world scenario not yet covered. While we observe improvement on multi-application tasks in OSWorld (3.8\% $\rightarrow$ 9.7\%), suggesting that single-website training already enables some cross-application transfer, generating coordinated multi-website environments with cross-site tasks remains a valuable direction for future work.

\paragraph{Mobile Evaluation.} While our generated websites use responsive layout design and we demonstrate positive transfer to MobileWorld (+3.9\%), our current system only generates web environments. Generating native mobile application environments for mobile GUI agent training is a direction for future work.

\paragraph{Generation Cost.} Generating a complete website environment requires multi-stage LLM calls, including task generation, architecture design, code generation, and test validation. The median generation time is approximately 20 minutes per website, at a cost of \$1.93 per site using GPT-5 batch pricing. While this cost is orders of magnitude lower than manual benchmark construction (e.g., OSWorld required approximately 1,800 person-hours for 369 tasks) and generation accounts for only 6.9\% of total training time due to embarrassingly parallel execution (600 websites in approximately 125 minutes), further reducing per-website cost and latency remains an engineering direction.

\bibliography{custom}

\newpage

\appendix

\section{Case Studies and Analysis}
\label{sec:appendix-case}

\subsection{Cross-Domain Transfer Analysis}
\label{sec:appendix-transfer}

\begin{figure*}[t]
\centering
\begin{subfigure}{0.48\textwidth}
  \centering
  \includegraphics[width=\textwidth]{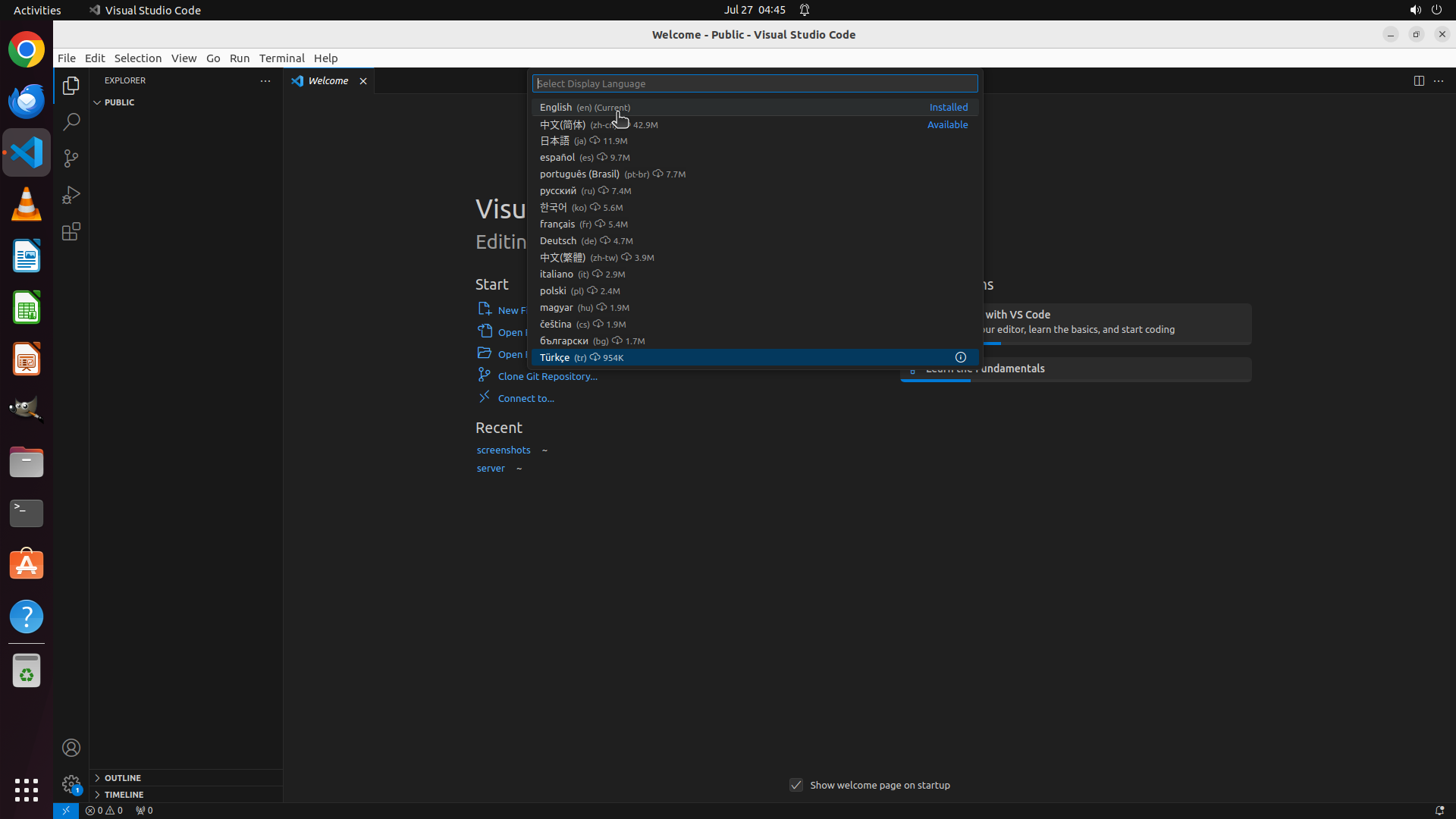}
  \caption{Baseline (Step 4): Language list visible, no Arabic found}
\end{subfigure}
\hfill
\begin{subfigure}{0.48\textwidth}
  \centering
  \includegraphics[width=\textwidth]{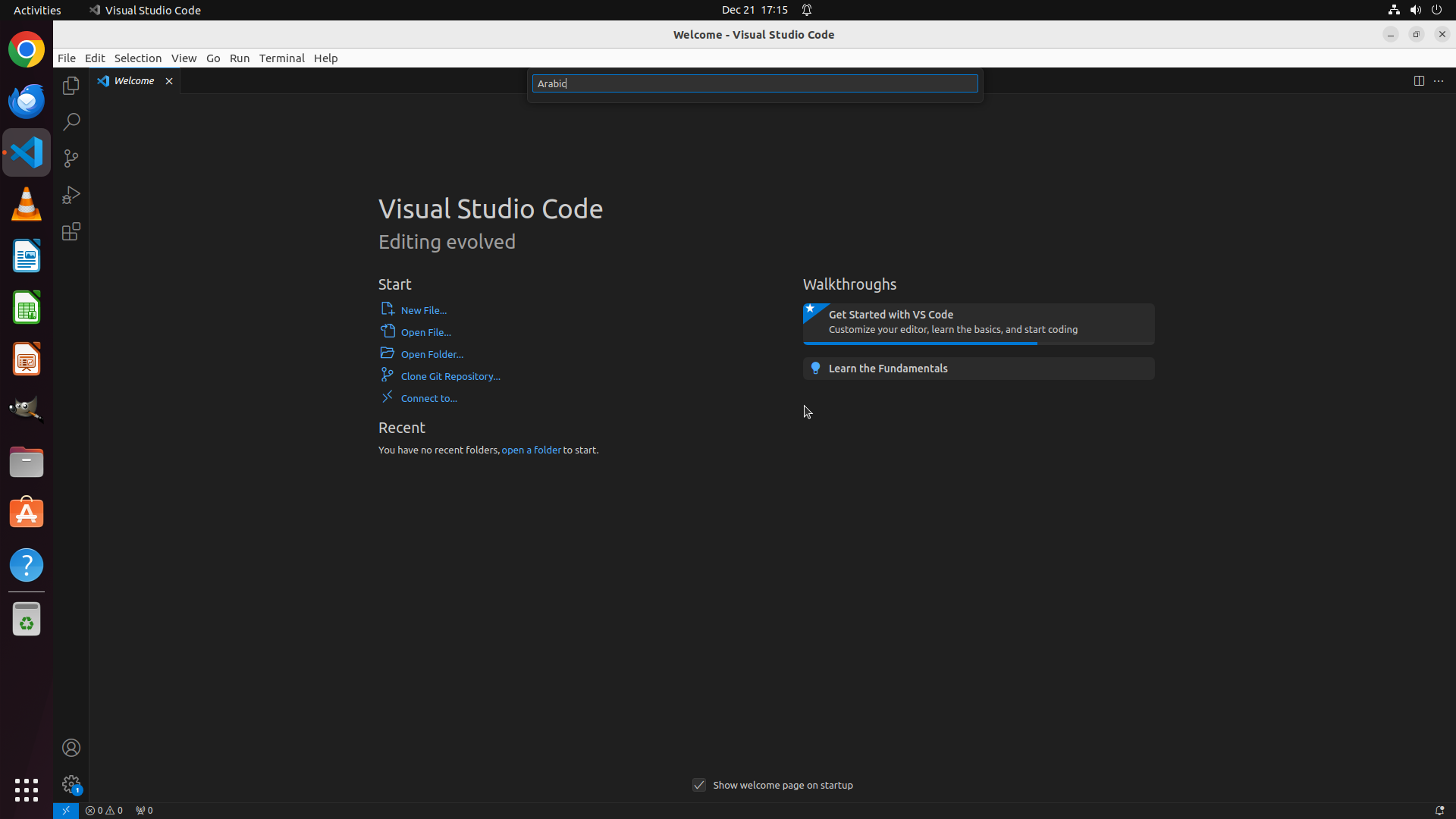}
  \caption{Trained (Step 5): Types ``Arabic'' to search}
\end{subfigure}

\vspace{0.8em}
\small
\begin{tabular}{c|p{5.5cm}|p{5.5cm}}
\toprule
\textbf{Step} & \textbf{Baseline} & \textbf{Trained} \\
\midrule
1--3 & Open command palette, search ``Configure Display Language'' & Same as baseline \\
4 & PageDown to browse list & Type ``Arabic'' in search \\
5 & \textbf{Give up}: ``Arabic not visible'' & Scroll, clear, retry \\
6--15 & -- & Continue exploring alternatives \\
\bottomrule
\end{tabular}
\caption{VS Code language change task: Exploration Persistence.}
\label{fig:transfer-vscode}
\end{figure*}

To understand why website training improves performance across all OSWorld domains, we analyzed execution traces from the multi-run experiments. To eliminate cases attributable to random variation, we applied strict filtering and focused on ``strong positive transfer'' cases, where the baseline failed consistently across all repeated runs while the trained model succeeded consistently. Analyzing the baseline failure patterns revealed three universal GUI interaction capabilities that website training develops:

\paragraph{Exploration Persistence.} The trained model persists in exploring alternatives when initial attempts fail, rather than prematurely giving up. In one VS Code task requiring language change to Arabic, the baseline browsed the language list with PageDown, concluded ``Arabic is not in the visible range,'' and terminated after only 5 steps. The trained model continued for 15 steps, trying multiple approaches (typing ``Arabic'', scrolling, clearing and retrying) until successfully locating and selecting the option. Figure~\ref{fig:transfer-vscode} shows the comparison.

\begin{figure*}[t]
\centering
\begin{subfigure}{0.32\textwidth}
  \centering
  \includegraphics[width=\textwidth]{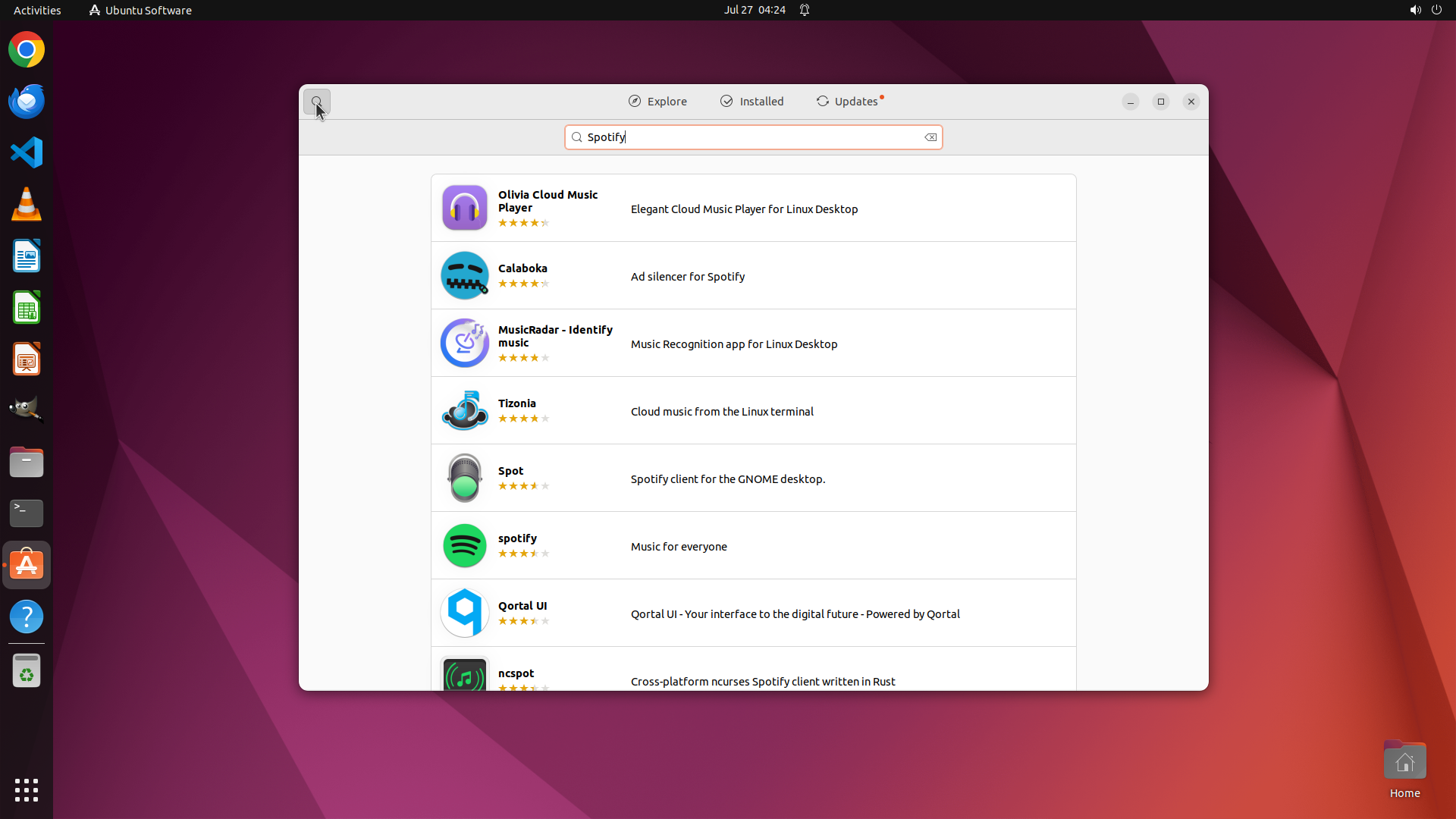}
  \caption{Baseline Step 4: Search results, task ends here}
\end{subfigure}
\hfill
\begin{subfigure}{0.32\textwidth}
  \centering
  \includegraphics[width=\textwidth]{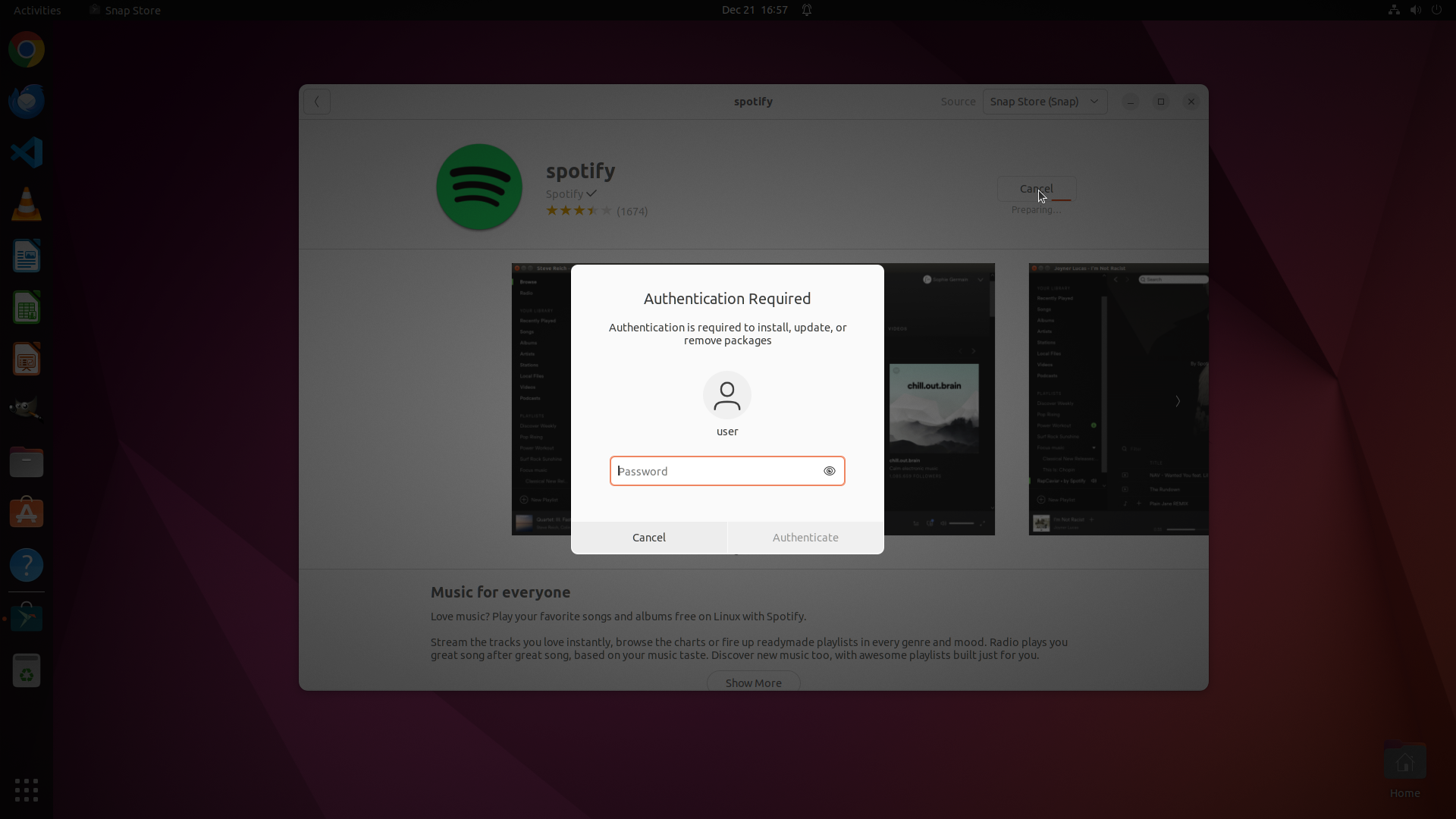}
  \caption{Trained Step 5: Auth dialog after Install}
\end{subfigure}
\hfill
\begin{subfigure}{0.32\textwidth}
  \centering
  \includegraphics[width=\textwidth]{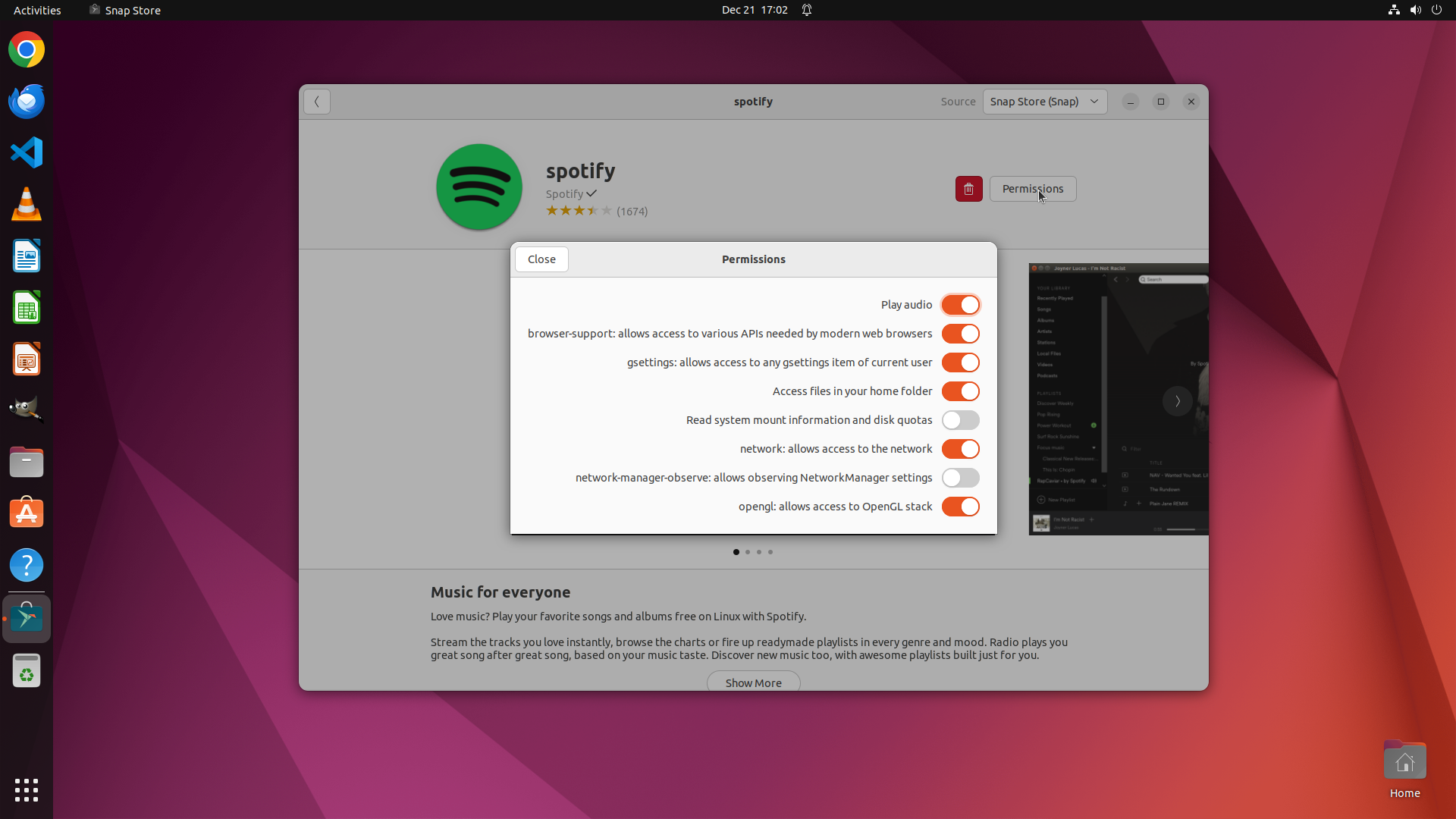}
  \caption{Trained Step 12: Configuring permissions}
\end{subfigure}

\vspace{0.8em}
\small
\begin{tabular}{c|p{5.5cm}|p{5.5cm}}
\toprule
\textbf{Step} & \textbf{Baseline} & \textbf{Trained} \\
\midrule
1--3 & Open Software Center, search ``Spotify'' & Same as baseline \\
4 & \textbf{Done}: Search results displayed & Click Spotify app \\
5 & -- & Click Install button \\
6--7 & -- & Enter password, authenticate \\
8--13 & -- & Wait for installation, configure permissions \\
\bottomrule
\end{tabular}
\caption{Spotify installation task: Flow Completeness. Baseline stops at search results; trained completes full installation.}
\label{fig:transfer-spotify}
\end{figure*}

\paragraph{Flow Completeness.} The trained model executes complete task workflows instead of stopping partway. For a Spotify installation task, the baseline opened Ubuntu Software Center, searched for ``Spotify,'' and called done after 4 steps, without clicking Install. The trained model completed the full 13-step flow: search, click Install, enter password for authentication, wait for installation progress, and verify completion. Figure~\ref{fig:transfer-spotify} illustrates this difference.

\begin{figure*}[t]
\centering
\begin{subfigure}{0.24\textwidth}
  \centering
  \includegraphics[width=\textwidth]{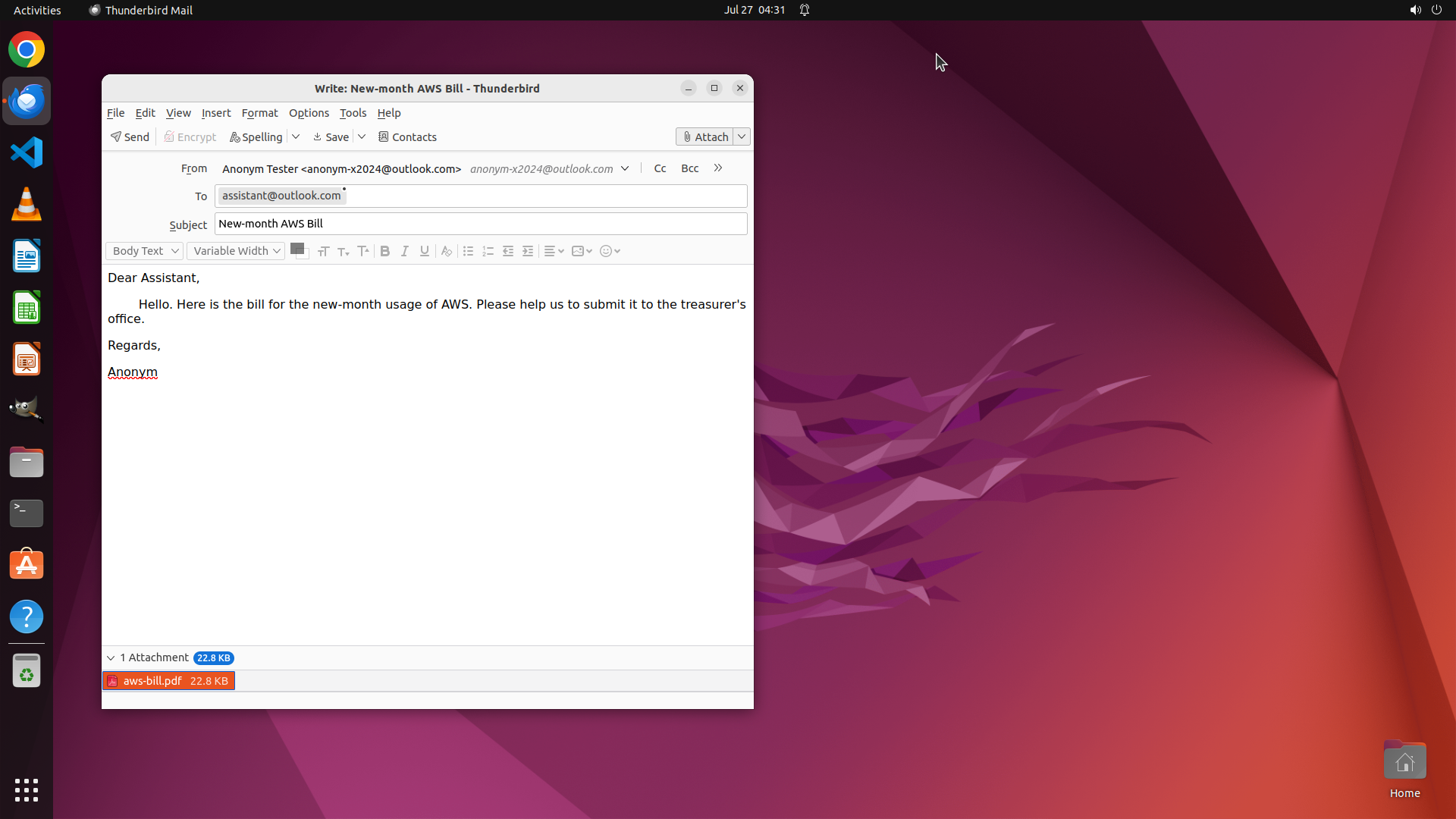}
  \caption{Baseline Step 4: Attachment added successfully}
\end{subfigure}
\hfill
\begin{subfigure}{0.24\textwidth}
  \centering
  \includegraphics[width=\textwidth]{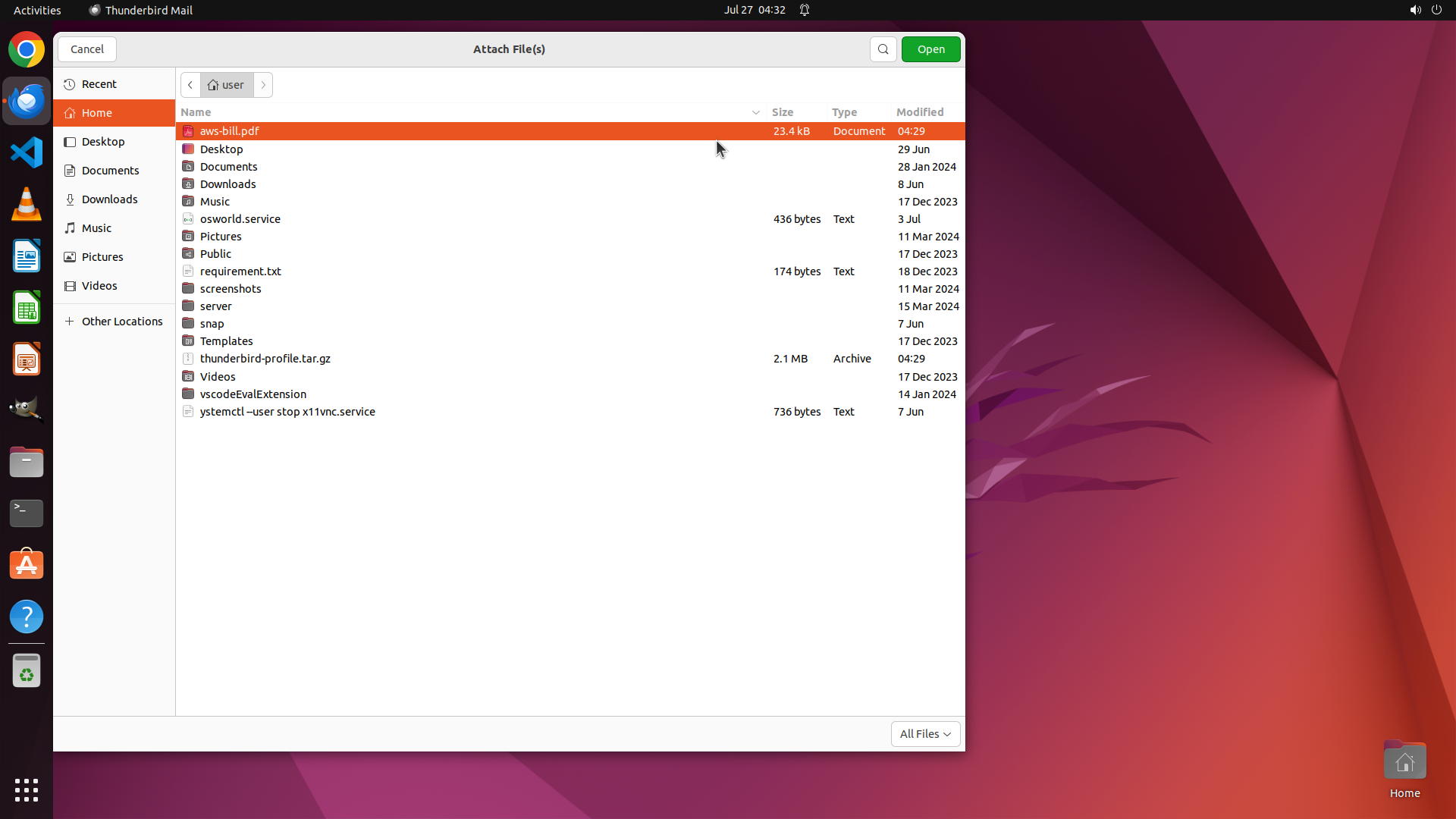}
  \caption{Baseline Step 5: File picker reopened}
\end{subfigure}
\hfill
\begin{subfigure}{0.24\textwidth}
  \centering
  \includegraphics[width=\textwidth]{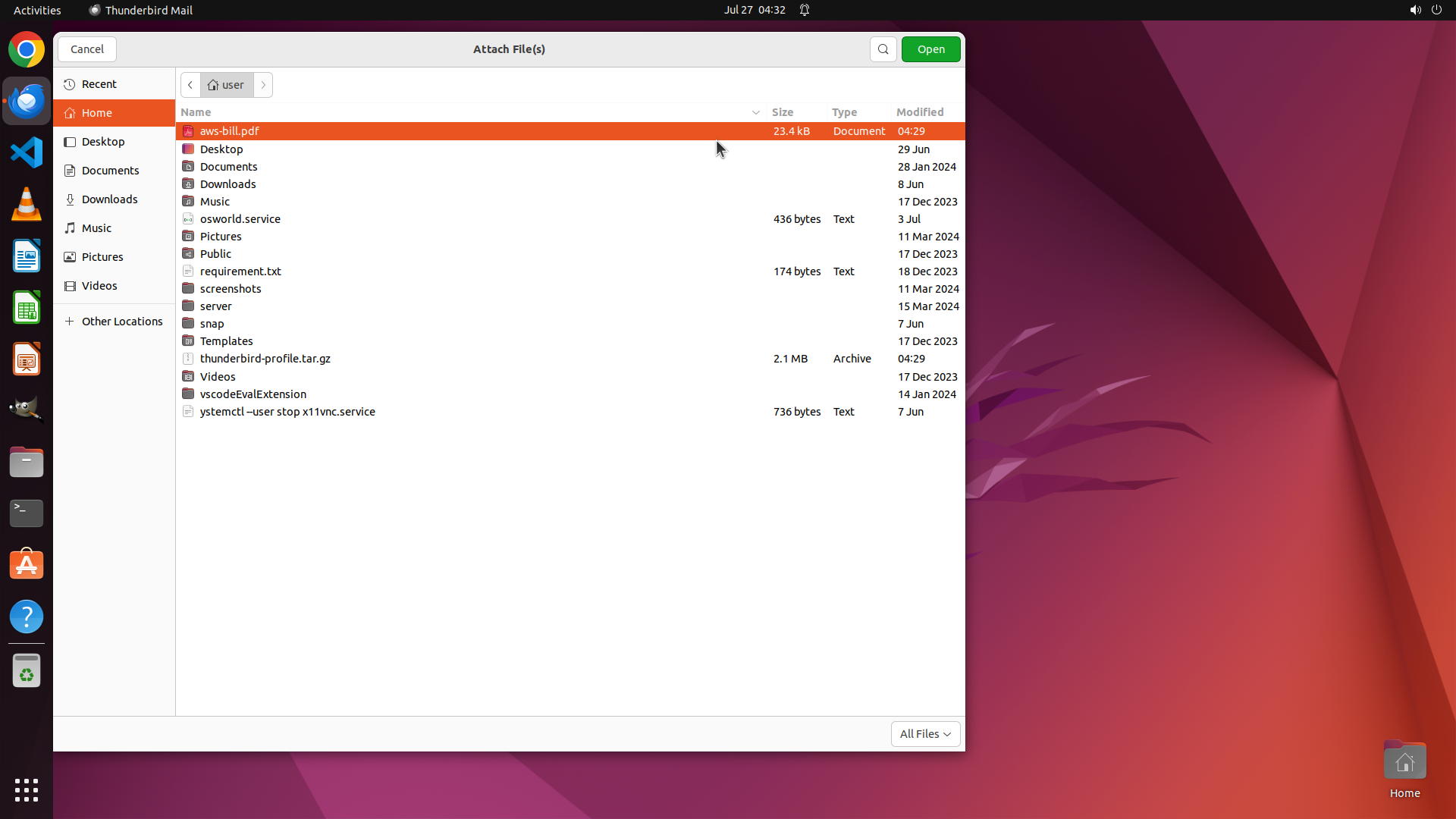}
  \caption{Baseline Step 7: Still in loop}
\end{subfigure}
\hfill
\begin{subfigure}{0.24\textwidth}
  \centering
  \includegraphics[width=\textwidth]{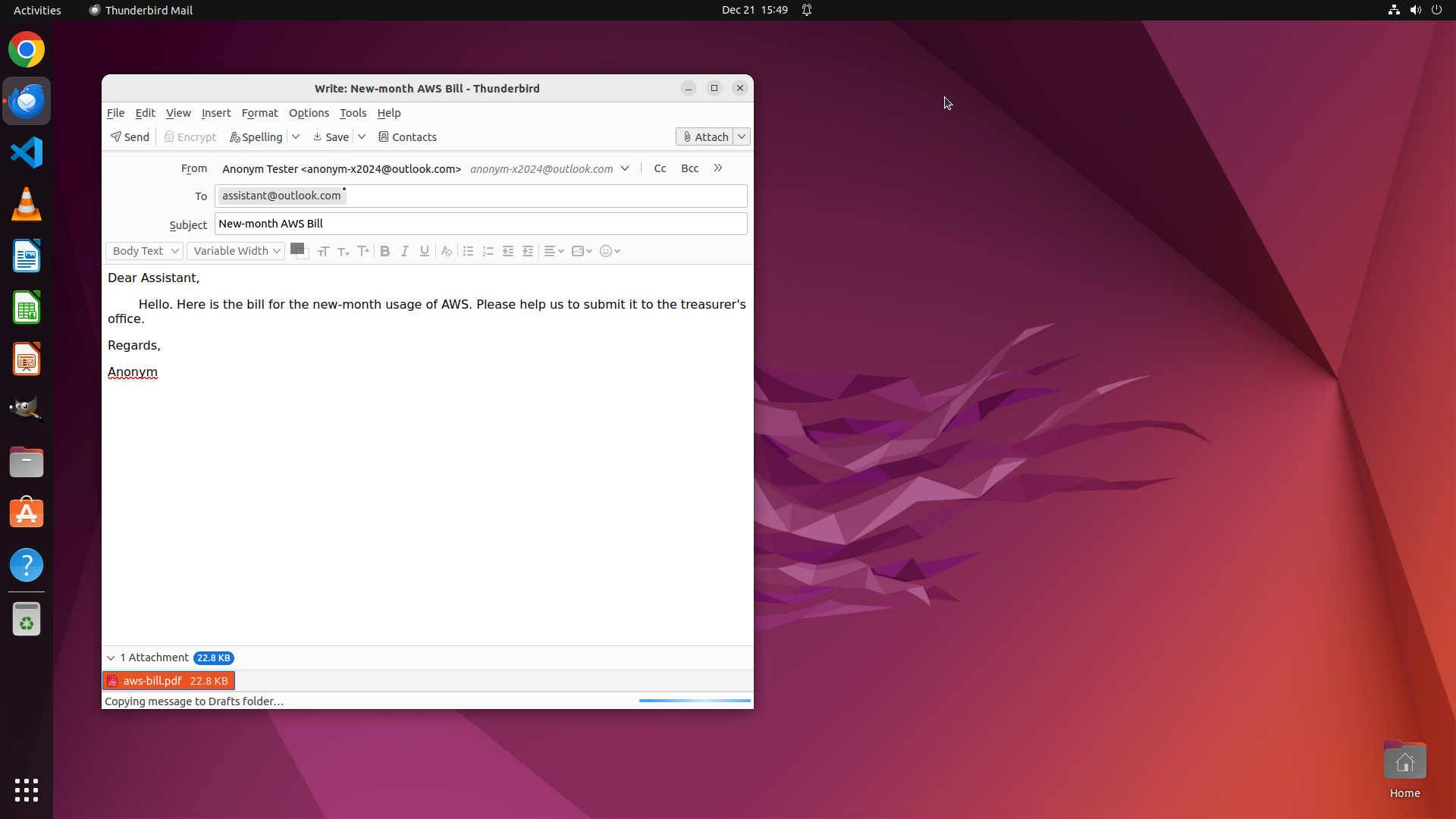}
  \caption{Trained Step 5: Task completed}
\end{subfigure}

\vspace{0.8em}
\small
\begin{tabular}{c|p{6cm}|p{5cm}}
\toprule
\textbf{Step} & \textbf{Baseline} & \textbf{Trained} \\
\midrule
1--4 & Attach file successfully (a--c) & Same as baseline \\
5 & \textbf{Loop starts}: Reopen file picker (b) & \textbf{Done}: Save to drafts (d) \\
6--15 & Repeat open/cancel 11 times (c shows step 7) & -- \\
\bottomrule
\end{tabular}
\caption{Thunderbird email attachment task: Loop Avoidance. After successfully attaching the file (a), baseline becomes confused and enters a futile loop (b, c), while trained model completes cleanly (d).}
\label{fig:transfer-thunderbird}
\end{figure*}

\paragraph{Loop Avoidance.} The trained model avoids getting stuck in repetitive action cycles. In an email attachment task in Thunderbird, the baseline successfully attached a file in 4 steps but then became confused about task completion, unsure what to do next after adding the attachment, it entered a futile loop of repeatedly opening the file picker and canceling for the remaining 11 steps. The trained model completed the same task cleanly in 5 steps. Figure~\ref{fig:transfer-thunderbird} demonstrates this pattern.

These capabilities are domain-agnostic: avoiding loops, completing workflows, and persisting through obstacles apply equally to image editors, office suites, and system utilities. Website environments, with their diverse interaction patterns and multi-step transactions, effectively train these transferable behaviors.

\subsection{Automatic Evaluator Generation}

InfiniteWeb automatically generates dense reward evaluators that provide proportional rewards for partial task completion. Figure~\ref{fig:evaluator-example} shows an evaluator for the task ``Subscribe to newsletter with weekly specials'' from a restaurant website.

\begin{figure*}[t]
\begin{tcolorbox}[title={\textbf{Generated Dense Reward Evaluator Example}}, colback=gray!5, colframe=black, colbacktitle=black, coltitle=white]
\small
\begin{verbatim}
const checkpoints = [];

// Load instrumentation and domain data
const attempt = JSON.parse(localStorage.getItem('task10_subscriptionAttempt'));
const confirmed = JSON.parse(localStorage.getItem('task10_subscriptionConfirmedOnSite'));
const subs = JSON.parse(localStorage.getItem('newslettersubscriptions') || '[]');

// CP1 (0.35): Subscription attempt with valid email and weekly specials enabled
const cp1 = attempt && isTestEmail(attempt.email) && attempt.wantsWeeklySpecials;
checkpoints.push({ passed: cp1, weight: 0.35 });

// CP2 (0.30): Subscription record matches attempt in storage
const subRecord = subs.find(s => s.id === attempt?.newsletterSubscriptionId);
const cp2 = subRecord && subRecord.email === attempt.email;
checkpoints.push({ passed: cp2, weight: 0.30 });

// CP3 (0.35): On-site confirmation references the subscription
const cp3 = confirmed?.newsletterSubscriptionId === attempt?.newsletterSubscriptionId;
checkpoints.push({ passed: cp3, weight: 0.35 });

// Dense reward: sum of weighted checkpoints (returns 0.0 to 1.0)
return checkpoints.reduce((sum, cp) => sum + (cp.passed ? cp.weight : 0), 0);
\end{verbatim}
\end{tcolorbox}
\caption{A generated dense reward evaluator with weighted checkpoints. Each checkpoint validates a different aspect: (1) user action tracking via instrumentation (weight 0.35), (2) data record consistency verification (weight 0.30), and (3) confirmation state validation (weight 0.35). Partial task completion yields proportional rewards, e.g., completing only the subscription attempt earns 0.35 points. This enables more effective GRPO training compared to sparse 0/1 rewards.}
\label{fig:evaluator-example}
\end{figure*}

The evaluator uses weighted checkpoints that enable dense reward signals for GRPO training. Each checkpoint validates a different aspect of task completion: (1) \textit{instrumentation flags} that track whether the agent performed required actions, (2) \textit{data consistency} that verifies records were properly created, and (3) \textit{confirmation state} that ensures the full workflow completed. The weighted sum allows partial credit, an agent that initiates but fails to complete a task still receives proportional reward. This design prevents shortcuts (directly manipulating localStorage fails instrumentation checks) while providing richer training signals than sparse 0/1 rewards.

\subsection{TCTDD Validation and Auto-Fix}

The TCTDD validation loop automatically detects and fixes implementation errors. Table~\ref{tab:tctdd-example} shows an example from a B2B industrial equipment website where one test initially failed.

\begin{table}[h]
\centering
\small
\begin{tabular}{p{0.95\columnwidth}}
\toprule
\textbf{Iteration 1: Test Execution} \\
\midrule
\texttt{$\checkmark$ Task 1: Request quote for three forklifts} \\
\texttt{$\checkmark$ Task 2: Add two electric forklifts to cart} \\
\texttt{$\times$ Task 3: Schedule demo for product} \\
\quad \textit{Error: Demo request should be submitted} \\
\texttt{$\checkmark$ Task 4--10: (passed)} \\
\textbf{Result: 9/10 passed, 1 failed} \\
\midrule
\textbf{Auto-Fix: LLM Analysis and Repair} \\
\midrule
The LLM identifies that \texttt{submitDemoRequest()} returns \texttt{undefined} instead of a success object, and generates a corrected implementation. \\
\midrule
\textbf{Iteration 2: Re-validation} \\
\midrule
\texttt{$\checkmark$ Task 1--10: All tests passed} \\
\textbf{Result: 10/10 passed} \\
\bottomrule
\end{tabular}
\caption{TCTDD validation loop example. The system detects a failing test, uses an LLM to analyze and fix the implementation, then re-validates until all tests pass.}
\label{tab:tctdd-example}
\end{table}

This iterative process ensures that the generated business logic correctly implements all required functionality. In our experiments, most websites require 1--3 iterations to pass all tests, with a maximum of 8 iterations allowed.

\section{Data Collection and Implementation}
\label{sec:appendix-impl}

\paragraph{Website Seed and Design Image Extraction.}
\label{sec:appendix-design-image}
We sample web pages from Common Crawl. For each sampled page, we render it in a headless browser and capture a full-page screenshot as the design image. We then use an LLM to analyze the visual content of the screenshot, generating a concise natural language description as the website seed, while filtering out pages that violate robots.txt or contain illegal content.

\paragraph{Generation Hyperparameters.} We use the following configuration: temperature 0.7, maximum output tokens 32,000, task count range 8--10 per website, maximum 12 pages per website, and maximum 8 iterations for TCTDD validation loop.

\paragraph{Agent Training.} We post-train UI-TARS-1.5-7B using GRPO. The training configuration includes: learning rate 1e-6, AdamW optimizer with bf16 precision, gradient clipping at 1.0, global batch size 16, PPO epochs 1, clip ratio 0.2--0.3, and discount factor $\gamma=0.95$. For rollout, we use 128 parallel environments, sample 8 trajectories per task, set maximum 15 steps per episode, and use temperature 1.0 for sampling.

\section{Experimental Details and Results}
\label{sec:appendix-webgenbench}

\paragraph{Baseline Implementation.} For the direct prompting baselines, we use GPT-5 with high reasoning effort as the backbone model. The prompt specifies website seed, required functionality, technical requirements (up to 12 pages, localStorage, reference design image), and code standards. The full prompt template is provided in Appendix~\ref{sec:appendix-prompts}.

\subsection{WebGen-Bench Results}
\label{sec:appendix-results}

Table~\ref{tab:webgen-detailed} presents the detailed results on WebGen-Bench across three independent runs. Welch's t-tests confirm that InfiniteWeb significantly outperforms all baselines: vs Bolt.diy ($t$=14.81, $p$<0.001), vs Claude-Code ($t$=6.33, $p$<0.01), and vs Codex ($t$=6.57, $p$<0.05).

Table~\ref{tab:webgen-ablation} shows the ablation study results. Both ablations show significant degradation: using GPT-4.1 instead of GPT-5 ($t$=7.70, $p$<0.01) and removing TCTDD validation ($t$=2.82, $p$<0.05).

\begin{table}[h]
\centering
\small
\begin{tabular}{lc}
\toprule
\textbf{Method} & \textbf{Task Completion (\%)} \\
\midrule
Bolt.diy & 67.1 $\pm$ 1.8 \\
Claude-Code & 75.8 $\pm$ 2.4 \\
Codex & 80.8 $\pm$ 0.4 \\
InfiniteWeb (Ours) & 85.6 $\pm$ 1.2 \\
\bottomrule
\end{tabular}
\caption{Detailed results on WebGen-Bench with standard deviation over three runs.}
\label{tab:webgen-detailed}
\end{table}

\begin{table}[h]
\centering
\small
\begin{tabular}{lc}
\toprule
\textbf{Configuration} & \textbf{Task Completion (\%)} \\
\midrule
GPT-4.1 (vs GPT-5) & 77.4 $\pm$ 1.4 \\
w/o TCTDD Validation & 82.6 $\pm$ 1.4 \\
InfiniteWeb (Full) & 85.6 $\pm$ 1.2 \\
\bottomrule
\end{tabular}
\caption{Ablation study results on WebGen-Bench.}
\label{tab:webgen-ablation}
\end{table}

\subsection{Online-Mind2Web Results}

Table~\ref{tab:mind2web-detailed} presents the results on Online-Mind2Web across three independent runs, broken down by task difficulty. Welch's t-tests comparing against baseline show: 200 tasks ($t$=2.96, $p$<0.05), 400 tasks ($t$=4.47, $p$<0.05), and 600 tasks ($t$=6.58, $p$<0.01).

\begin{table}[h]
\centering
\small
\begin{tabular}{lcccc}
\toprule
\textbf{Training} & \textbf{Easy} & \textbf{Medium} & \textbf{Hard} & \textbf{Overall} \\
\midrule
Orig. & 46.9 & 18.9 & 5.3 & \textbf{23.0$\pm$0.9} \\
200 & 54.3 & 18.2 & 7.9 & \textbf{25.3$\pm$1.0} \\
400 & 48.2 & 27.3 & 5.3 & \textbf{27.3$\pm$1.4} \\
600 & 56.8 & 23.1 & 9.2 & \textbf{28.7$\pm$1.2} \\
\bottomrule
\end{tabular}
\caption{Results on Online-Mind2Web by difficulty (\%). 600/400/200 denote InfiniteWeb with different training task counts. Orig. is the original baseline. Standard deviation computed over three runs.}
\label{tab:mind2web-detailed}
\end{table}

\subsection{Appearance Win Rate}

Table~\ref{tab:appearance-detailed} presents the appearance comparison results across three independent runs.

\begin{table}[h]
\centering
\small
\begin{tabular}{lc}
\toprule
\textbf{Comparison} & \textbf{Win Rate (\%)} \\
\midrule
Ours vs Claude-Code & 85.4 $\pm$ 0.5 \\
Ours vs Bolt.diy & 84.2 $\pm$ 0.8 \\
Ours vs Codex & 69.3 $\pm$ 0.7 \\
\bottomrule
\end{tabular}
\caption{Appearance win rate comparison. Win rate indicates how often InfiniteWeb-generated websites are judged visually closer to the reference design image. Standard deviation computed over three runs.}
\label{tab:appearance-detailed}
\end{table}

\subsection{Comparison with Claude-Code-Generated Websites}
\label{sec:appendix-claude-code}

To disentangle the effect of data quality from data quantity, we compare training on InfiniteWeb-generated websites against training on Claude-Code-generated websites using the same 600-task GRPO setup. Both use GPT-5 with reasoning effort set to ``high'' for generation. All results are averaged over three runs.

\begin{table}[h]
\centering
\small
\resizebox{\columnwidth}{!}{%
\begin{tabular}{lccc}
\toprule
\textbf{Training Data} & \textbf{Mind2Web} & \textbf{OSWorld} & \textbf{MobileWorld} \\
\midrule
Baseline & 23.0 & 24.5 & 6.4 \\
Claude-Code & 23.7 {\scriptsize(+0.7)} & 25.2 {\scriptsize(+0.7)} & 5.1 {\scriptsize(-1.3)} \\
InfiniteWeb & \textbf{28.7} {\scriptsize(+5.7)} & \textbf{31.4} {\scriptsize(+6.9)} & \textbf{10.3} {\scriptsize(+3.9)} \\
\bottomrule
\end{tabular}}
\caption{Comparison of training on InfiniteWeb vs.\ Claude-Code-generated websites (\%). Both use 600 tasks with identical GRPO training. Mind2Web refers to Online-Mind2Web.}
\label{tab:claude-code-comparison}
\end{table}

Training on Claude-Code-generated websites yields marginal or even negative improvements, while InfiniteWeb achieves significant gains across all three benchmarks. This demonstrates that data quality, not quantity, drives the improvement. The quality gap stems from context management: InfiniteWeb provides each generation step with only directly relevant specifications (interface definitions, data models), while Claude-Code accumulates the full codebase in context, leading to cross-page inconsistencies, reduced scalability (4.9 vs.\ 12.6 pages per website), and less complete business logic as the context grows.

\section{Human Evaluation for Visual Quality}
\label{sec:appendix-human}

To validate the reliability of our automated visual evaluation (Section~\ref{sec:experiments}), we conducted a human verification study. We randomly sampled 100 comparison cases across all three baseline comparisons (InfiniteWeb-Codex, InfiniteWeb-Claude, and InfiniteWeb-Bolt), with approximately equal representation from each. Human evaluators were presented with the reference design image and two website screenshots (A and B), and asked to determine which implementation more closely matches the reference design.

The human judgments achieved a 91\% agreement rate with the automated GPT-5 evaluations, indicating that the automated visual quality assessment is highly reliable and well-aligned with human perception. The disagreement cases primarily involved subtle differences where both implementations were reasonably close to the reference design, making the distinction less clear-cut.

\section{Human Verification of Task and Evaluator Quality}
\label{sec:appendix-task-quality}

To validate the quality of generated tasks and automatic evaluators, we conducted a manual verification study. We randomly sampled 100 tasks from the generated websites and had human evaluators assess: (1) whether the task description is clear and executable on the generated website, and (2) whether the automatic evaluator correctly determines task completion.

Of the 100 sampled tasks, 95 passed human verification, confirming that our system generates high-quality tasks with reliable automatic evaluators.

Additionally, we analyzed the TCTDD validation loop statistics. Among all generated websites, only 1.5\% remained unfixed after the maximum 8 TCTDD iterations, demonstrating the effectiveness of our iterative test-driven approach in ensuring functional correctness.

\section{Prompts}
\label{sec:appendix-prompts}

\begin{figure*}[t]
\begin{tcolorbox}[title={\textbf{Prompt Template for Baseline Website Generation}}, colback=gray!5, colframe=black, colbacktitle=black, coltitle=white]
\small
Generate a full static \{website\_type\} website using HTML/CSS/JavaScript in the current directory.

\textbf{BASIC REQUIREMENTS:}
\begin{itemize}[nosep,leftmargin=*]
\item Create up to 12 pages
\item Homepage must be index.html
\item Use browser's localStorage to store data
\item Use design\_image.png as the visual design reference
\item The website must implement at least these functions: \{function\_requirements\}
\end{itemize}

\textbf{FILE STRUCTURE:}
\begin{itemize}[nosep,leftmargin=*]
\item Each HTML page should have its own CSS file (e.g., index.html $\rightarrow$ index.css)
\end{itemize}

\textbf{DESIGN MATCHING:}
\begin{itemize}[nosep,leftmargin=*]
\item Carefully analyze design\_image.png and extract: color scheme, typography, layout patterns, spacing system
\item Accurately reproduce the visual style to ensure design consistency across all pages
\end{itemize}

\textbf{HTML STANDARDS:}
\begin{itemize}[nosep,leftmargin=*]
\item Use semantic HTML5 elements: \texttt{<header>}, \texttt{<main>}, \texttt{<footer>}, \texttt{<nav>}, \texttt{<section>}, \texttt{<article>}
\end{itemize}

\textbf{CSS STANDARDS:}
\begin{itemize}[nosep,leftmargin=*]
\item Use modern CSS features: Flexbox, Grid, CSS Custom Properties
\item Implement interactive states: hover effects and smooth transitions
\end{itemize}

\textbf{JAVASCRIPT STANDARDS:}
\begin{itemize}[nosep,leftmargin=*]
\item Use modern ES6+ syntax: template literals, arrow functions, const/let, destructuring
\end{itemize}

\textbf{FUNCTIONALITY REQUIREMENTS:}
\begin{itemize}[nosep,leftmargin=*]
\item Ensure all specified user tasks can be completed end-to-end
\item Fully implement each page's core functionality, not just static displays
\item Beyond the explicitly required functions, add other common features appropriate for this website seed
\item When page parameters are missing, provide reasonable default content
\end{itemize}

\textbf{DATA QUALITY:}
\begin{itemize}[nosep,leftmargin=*]
\item Ensure temporal data alignment: dates should be logically consistent
\item Generate diverse data with sufficient variety to support different scenarios
\item Create realistic, professional content appropriate for the website seed
\end{itemize}
\end{tcolorbox}
\caption{Prompt template for baseline website generation. Variables \{website\_type\} and \{function\_requirements\} are filled based on the input specification.}
\label{fig:baseline-prompt}
\end{figure*}

\begin{figure*}[t]
\begin{tcolorbox}[title={\textbf{Prompt: Task Generation}}, colback=gray!5, colframe=black, colbacktitle=black, coltitle=white]
\small
You are a UX researcher. Generate \{task\_count\_range\} realistic user tasks for a \{website\_type\}.

\textbf{IMPORTANT REQUIREMENTS:}
\begin{enumerate}[nosep,leftmargin=*]
\item This is a mock website, so tasks should NOT depend on any external services like email authentication.
\item Each task MUST contain between \{min\_steps\}-\{max\_steps\} detailed steps for proper complexity.
\item Tasks should be suitable for RL model training, requiring multiple decisions and interactions.
\end{enumerate}

\textbf{Each task should:}
\begin{itemize}[nosep,leftmargin=*]
\item Represent a SPECIFIC user goal with MEASURABLE success criteria
\item Contain \{min\_steps\}-\{max\_steps\} DETAILED action steps
\item Include CLEAR decision criteria (e.g., ``select the cheapest option'', ``choose items with 4+ stars'')
\item Specify EXACT targets (e.g., ``add 3 items under \$50'', ``find products with free shipping'')
\item Use CONCRETE values and thresholds (prices, quantities, ratings, dates)
\item Cover different aspects of the website functionality
\end{itemize}

\textbf{Task specificity requirements:}
\begin{itemize}[nosep,leftmargin=*]
\item BAD: ``Compare products and select the best one''
\item GOOD: ``Compare two laptops and select the one with more RAM under \$1000''
\item BAD: ``Search for headphones and add to cart''
\item GOOD: ``Search for wireless headphones under \$200 with 4+ star rating and add the first result to cart''
\end{itemize}

\textbf{Step detail requirements (FOCUS ON ACTIONS, NOT VERIFICATION):}
\begin{itemize}[nosep,leftmargin=*]
\item Specific navigation actions (e.g., ``Navigate to the homepage'')
\item Clear element interactions (e.g., ``Click the search button in the header'')
\item Precise data entry (e.g., ``Type `wireless headphones' in the search field'')
\item Selection actions (e.g., ``Select `Blue' from the color dropdown'')
\item Page transitions (e.g., ``Click on the product image to open details page'')
\end{itemize}

\textbf{AVOID these types of steps:}
\begin{itemize}[nosep,leftmargin=*]
\item Verification steps (e.g., ``Verify the page loaded'')
\item Validation steps (e.g., ``Validate the price is correct'')
\item Confirmation steps (e.g., ``Ensure the button is visible'')
\end{itemize}

\textbf{Return JSON format:}
\begin{verbatim}
{"tasks": [{"id": "task_1", "name": "...",
  "description": "...", "steps": ["..."]}]}
\end{verbatim}
\end{tcolorbox}
\caption{Prompt for automatic task generation from website seed.}
\label{fig:prompt-task-generation}
\end{figure*}

\begin{figure*}[t]
\begin{tcolorbox}[title={\textbf{Prompt: Primary Architecture Design}}, colback=gray!5, colframe=black, colbacktitle=black, coltitle=white]
\small
Design a complete website architecture for a \{website\_type\}.

\textbf{User Tasks that the website must support:}

\{tasks\_text\}

Based on these tasks, design a COMPLETE architecture with ALL pages needed:
\begin{enumerate}[nosep,leftmargin=*]
\item All pages needed for the website (maximum \{max\_pages\} pages)
\item Primary functions each page should provide
\item Keep it simple and focused on user needs
\item DO NOT include authentication/login pages
\item DO NOT consider multi-user scenarios
\item This is for single-user use only
\end{enumerate}

\textbf{Return JSON format:}
\begin{verbatim}
{"all_pages": [{"name": "Page Name", "filename": "page.html"}],
 "pages": [{"name": "Page Name", "filename": "page.html",
   "description": "Brief description",
   "primary_functions": ["Function 1", "Function 2"]}]}
\end{verbatim}

\textbf{Requirements:}
\begin{itemize}[nosep,leftmargin=*]
\item Include all pages needed to complete the user tasks
\item Each page should have clear, focused responsibilities
\item Use descriptive filenames (e.g., index.html, products.html, cart.html)
\item Primary functions should be high-level user actions
\item Ensure all task steps can be completed with the designed pages
\item index.html must be contained and as the homepage
\end{itemize}
\end{tcolorbox}
\caption{Prompt for designing primary website architecture based on user tasks.}
\label{fig:prompt-primary-architecture}
\end{figure*}

\begin{figure*}[t]
\begin{tcolorbox}[title={\textbf{Prompt: Data Model Extraction}}, colback=gray!5, colframe=black, colbacktitle=black, coltitle=white]
\small
You are a data architect. Analyze the user tasks and extract ALL data entities and fields needed.

\textbf{Website Seed:} \{website\_seed\}

\textbf{User Tasks:} \{tasks\_json\}

\textbf{Website Architecture Pages:} \{pages\_json\}

For each task, identify:
\begin{enumerate}[nosep,leftmargin=*]
\item Core entities directly mentioned (e.g., Product, Cart)
\item Supporting entities needed for functionality
\item All necessary fields for each entity
\item Relationships between entities
\end{enumerate}

\textbf{IMPORTANT REQUIREMENTS:}
\begin{itemize}[nosep,leftmargin=*]
\item This is for SINGLE-USER agent training only - NO multi-user support needed
\item DO NOT include User entity or userId/sessionId fields
\item DO NOT include authentication-related entities
\item Extract ALL entities needed, not just the minimal set
\item Include all fields necessary for the tasks
\item Specify data types for each field
\item Identify primary keys (but NO foreign keys to User)
\item Specify data\_pre\_generation\_num for each entity: ``many'', ``few'', or ``none''
  \begin{itemize}[nosep,leftmargin=*]
  \item ``many'': Generate 10-20 items (for catalog entities like Product, Category)
  \item ``few'': Generate 3-5 items (for limited entities like Brand, Department)
  \item ``none'': No pre-generation needed (for runtime entities like Cart, Order)
  \end{itemize}
\item Provide storage\_key for localStorage (lowercase plural form)
\end{itemize}

\textbf{Return JSON format:}
\begin{verbatim}
{"entities": [{"name": "Product", "storage_key": "products",
  "fields": [{"name": "id", "type": "string", "primary_key": true},
   {"name": "price", "type": "number", "required": true}],
  "data_pre_generation_num": "many"}],
 "relationships": [{"from": "CartItem", "to": "Product",
  "type": "belongs_to", "field": "productId"}]}
\end{verbatim}
\end{tcolorbox}
\caption{Prompt for extracting data models from user tasks.}
\label{fig:prompt-data-extraction}
\end{figure*}

\begin{figure*}[t]
\begin{tcolorbox}[title={\textbf{Prompt: Interface Design}}, colback=gray!5, colframe=black, colbacktitle=black, coltitle=white]
\small
You are a software architect. Design comprehensive interfaces for both user tasks AND page functionality.

\textbf{Website Seed:} \{website\_seed\}

\textbf{User Tasks:} \{tasks\_json\}

\textbf{Data Models:} \{data\_models\_json\}

\textbf{Website Pages and Functions:} \{pages\_info\}

\textbf{IMPORTANT REQUIREMENTS:}
\begin{enumerate}[nosep,leftmargin=*]
\item Design USER-FACING interfaces that will be directly called from UI
\item This is for SINGLE-USER agent training - NO userId, sessionId parameters
\item System state (cart, session) should be managed internally, not passed as parameters
\end{enumerate}

\textbf{CRITICAL: Design interfaces for TWO purposes:}

\textbf{A. TASK EXECUTION INTERFACES} - For user tasks:
\begin{itemize}[nosep,leftmargin=*]
\item What information must be shown BEFORE the user can act (display interfaces)
\item What action the user performs (action interfaces)
\item What feedback/results need to be shown AFTER the action (result interfaces)
\end{itemize}

\textbf{B. PAGE FUNCTIONALITY INTERFACES} - For each page's primary\_functions:
\begin{itemize}[nosep,leftmargin=*]
\item Review EVERY primary\_function in the Website Pages list
\item Ensure there's an interface to support EACH function
\item Examples: ``Navigate to featured product categories'' $\rightarrow$ needs getCategories()
\end{itemize}

\textbf{Additional requirements:}
\begin{itemize}[nosep,leftmargin=*]
\item Interfaces should handle complete operations (e.g., addToCart handles cart creation if needed)
\item Do NOT create unnecessary CRUD, but DO create interfaces needed for page display
\item For interfaces that get data for display, return user-friendly fields
\end{itemize}

\textbf{Return JSON format:}
\begin{verbatim}
{"interfaces": [{"name": "addToCart",
  "description": "Add a product to cart",
  "parameters": [{"name": "productId", "type": "string"}],
  "returns": {"type": "object",
    "properties": {"success": {"type": "boolean"}}},
  "relatedTasks": ["task_1"]}],
 "helperFunctions": [{"name": "_getOrCreateCart",
  "description": "Internal helper", "visibility": "private"}]}
\end{verbatim}
\end{tcolorbox}
\caption{Prompt for designing user-facing interfaces based on tasks and data models.}
\label{fig:prompt-interface-design}
\end{figure*}

\begin{figure*}[t]
\begin{tcolorbox}[title={\textbf{Prompt: Interface Wrapping}}, colback=gray!5, colframe=black, colbacktitle=black, coltitle=white]
\small
You are a software architect analyzing interface parameters for a \{website\_type\}.

Your task: Identify parameters that should be hidden from user-facing interfaces and generate wrapped versions.

\textbf{ORIGINAL INTERFACES:} \{original\_interfaces\_json\}

\textbf{EXISTING DATA MODELS:} \{data\_models\_json\}

\textbf{PARAMETER CLASSIFICATION RULES:}

\textbf{1. SYSTEM-MANAGED PARAMETERS (should be hidden):}
\begin{itemize}[nosep,leftmargin=*]
\item User identity: userId, guestId, sessionId, currentUser
\item System context: cartId, deviceId, timestamp, requestId
\item Authentication: authToken, userRole, permissions, isAuthenticated
\item Environment: locale, timezone, region, currency
\end{itemize}

\textbf{2. USER-PROVIDED PARAMETERS (should remain exposed):}
\begin{itemize}[nosep,leftmargin=*]
\item Business data: productId, quantity, rating, comment
\item User selections: selectedSize, color, filters
\item User input: searchQuery, address, paymentDetails
\end{itemize}

\textbf{ANALYSIS CRITERIA:}
\begin{itemize}[nosep,leftmargin=*]
\item Ask: ``Would a user type this value into a form or select it from a UI?''
\item If YES $\rightarrow$ Keep as parameter (user-provided)
\item If NO $\rightarrow$ Hide and manage through state (system-managed)
\end{itemize}

\textbf{EXAMPLE TRANSFORMATION:}

Original: addToCart(userId, guestId, productId, quantity, selectedSize)

Wrapped: addToCart(productId, quantity, selectedSize)

State Needed: UserSession with currentUserId/currentGuestId

\textbf{Return JSON format:}
\begin{verbatim}
{"wrapped_interfaces": [{"name": "addToCart",
  "parameters": [{"name": "productId", "type": "string"}]}],
 "state_data_models": [{"name": "UserSession",
  "fields": [{"name": "currentUserId", "type": "string"}]}],
 "implementation_mapping": [{"wrapped_function": "addToCart",
  "parameter_mapping": {"userId": "_getSession().currentUserId"}}]}
\end{verbatim}
\end{tcolorbox}
\caption{Prompt for wrapping interfaces to hide system-managed parameters.}
\label{fig:prompt-interface-wrapping}
\end{figure*}

\begin{figure*}[t]
\begin{tcolorbox}[title={\textbf{Prompt: Architecture Design}}, colback=gray!5, colframe=black, colbacktitle=black, coltitle=white]
\small
You are a web architect. Design complete website architecture based on user tasks and interfaces.

\textbf{Website Seed:} \{website\_seed\}

\textbf{User Tasks:} \{task\_summary\_json\}

\textbf{Primary Architecture (initial design):} \{primary\_arch\_json\}

\textbf{Available Interfaces:} \{interface\_summary\_json\}

\textbf{Data Models:} \{data\_summary\_json\}

\textbf{IMPORTANT:}
\begin{itemize}[nosep,leftmargin=*]
\item This is for SINGLE-USER agent training - NO authentication/login pages needed
\item The interfaces provided are USER-FACING interfaces (no userId/sessionId parameters)
\item System state is managed automatically through localStorage
\end{itemize}

\textbf{Design Requirements:}
\begin{enumerate}[nosep,leftmargin=*]
\item Use EXACTLY the pages from primary architecture - do not add or remove pages
\item Assign appropriate interfaces to each page based on functionality
\item Use URL parameters for navigation (NOT localStorage for page data)
\item Define incoming parameters (what parameters the page accepts)
\item Define outgoing connections (what pages this page navigates to)
\item Specify access methods for each page
\item Design header and footer navigation links
\end{enumerate}

\textbf{Access Method Guidelines:}
\begin{itemize}[nosep,leftmargin=*]
\item ``navigation'': Accessible through header/footer navigation
\item ``url\_param'': Accessible through URL parameters from other pages
\item ``direct\_link'': Accessible through direct links in content
\item ``form\_submission'': Accessible after form submission
\end{itemize}

\textbf{Return JSON format:}
\begin{verbatim}
{"all_pages": [{"name": "Home", "filename": "index.html"}],
 "pages": [{"name": "Home", "filename": "index.html",
   "assigned_interfaces": ["searchProducts"],
   "incoming_params": [],
   "outgoing_connections": [{"target": "product.html",
     "params": {"id": "productId"}}],
   "access_methods": [{"type": "navigation"}]}],
 "header_links": [{"text": "Home", "url": "index.html"}]}
\end{verbatim}
\end{tcolorbox}
\caption{Prompt for designing complete website architecture with page navigation.}
\label{fig:prompt-architecture-design}
\end{figure*}

\begin{figure*}[t]
\begin{tcolorbox}[title={\textbf{Prompt: Page Functionality Design}}, colback=gray!5, colframe=black, colbacktitle=black, coltitle=white]
\small
You are a senior web functional designer. Design the functional aspects and workflows of a webpage.

\textbf{Website Seed:} \{website\_seed\}

\textbf{Page Architecture:} \{page\_spec\_json\}

\textbf{Available Data Models:} \{data\_dict\_json\}

\textbf{Assigned Interfaces for This Page:} \{interface\_details\_json\}

\textbf{Navigation Information:} \{navigation\_info\}

\textbf{DESIGN REQUIREMENTS:}
\begin{enumerate}[nosep,leftmargin=*]
\item Create an engaging, specific page title
\item Write a rich, detailed description of the page
\item Design core features based on the assigned interfaces
\item Define user workflows that utilize the interfaces
\item Specify user interactions (clicks, forms, navigation)
\item Describe state logic using URL parameters (NOT localStorage)
\item Create functional components that use the interfaces
\end{enumerate}

\textbf{IMPORTANT GUIDELINES:}
\begin{itemize}[nosep,leftmargin=*]
\item Use ONLY the assigned interfaces for this page
\item Navigation uses URL parameters (e.g., product.html?id=123)
\item Focus on functionality, not visual appearance
\item Components should be functional, not presentational
\item Each component should have clear data binding and event handlers
\item Output should not involve any static data or hardcoded values
\end{itemize}

\textbf{Return JSON format:}
\begin{verbatim}
{"title": "Page title", "description": "Page description",
 "page_functionality": {"core_features": ["Feature 1"],
   "user_workflows": ["Workflow step"],
   "interactions": ["Click action"],
   "state_logic": "URL parameter handling"},
 "components": [{"id": "search-form", "type": "search-form",
   "functionality": "Handles product search",
   "data_binding": ["Product"],
   "event_handlers": ["onSubmit"]}]}
\end{verbatim}
\end{tcolorbox}
\caption{Prompt for designing page functionality and components.}
\label{fig:prompt-page-design}
\end{figure*}

\begin{figure*}[t]
\begin{tcolorbox}[title={\textbf{Prompt: Design Image Analysis}}, colback=gray!5, colframe=black, colbacktitle=black, coltitle=white]
\small
You are a senior UI/UX design analyst. Analyze the provided design image to extract all visual characteristics.

\textbf{Website Seed:} \{website\_seed\}

\textbf{ANALYSIS TASKS:}

\textbf{1. Visual Features Analysis:}
\begin{itemize}[nosep,leftmargin=*]
\item Identify overall visual style (modern, minimalist, vibrant, corporate, etc.)
\item Describe visual hierarchy and focal points
\item Note use of whitespace and visual breathing room
\end{itemize}

\textbf{2. Color Scheme Extraction:}
\begin{itemize}[nosep,leftmargin=*]
\item Primary colors (main brand colors)
\item Secondary colors (supporting colors)
\item Accent colors (for CTAs, highlights)
\item Neutral colors (backgrounds, text, borders)
\item Provide exact hex color values
\end{itemize}

\textbf{3. Layout Characteristics:}
\begin{itemize}[nosep,leftmargin=*]
\item Grid system (12-column, custom, etc.)
\item Layout patterns (sidebar, centered, full-width)
\item Section organization and alignment principles
\end{itemize}

\textbf{4. UI Patterns:} Button styles, card designs, form elements, navigation patterns

\textbf{5. Typography:} Font families, size hierarchy, font weights, line heights

\textbf{6. Spacing System:} Base unit, padding/margin patterns, component spacing

\textbf{7. Interaction Hints:} Hover states, transitions, animation suggestions

\textbf{Return JSON format:}
\begin{verbatim}
{"visual_features": {"overall_style": "modern minimalist"},
 "color_scheme": {"primary": ["#hex"], "accent": ["#hex"]},
 "layout_characteristics": {"grid_system": "12-column"},
 "ui_patterns": [{"pattern_type": "button",
   "characteristics": {"shape": "rounded"}}],
 "typography": {"font_families": {"heading": "Inter"}},
 "spacing_system": {"base_unit": "8px"}}
\end{verbatim}
\end{tcolorbox}
\caption{Prompt for analyzing design image to extract visual characteristics.}
\label{fig:prompt-design-analysis}
\end{figure*}

\begin{figure*}[t]
\begin{tcolorbox}[title={\textbf{Prompt: Layout Design}}, colback=gray!5, colframe=black, colbacktitle=black, coltitle=white]
\small
You are a senior UI/UX designer. Create a thoughtful, detailed layout for existing components.

\textbf{DESIGN DNA (extracted from design image):}
\begin{itemize}[nosep,leftmargin=*]
\item Visual Style: \{visual\_style\}
\item Grid System: \{grid\_system\}
\item Layout Pattern: \{layout\_pattern\}
\item Spacing System: \{spacing\_system\_json\}
\end{itemize}

\textbf{PAGE CONTEXT:} Website Seed: \{website\_seed\}, Page: \{page\_name\}

\textbf{Components to Layout:} \{components\_list\}

\textbf{STEP 1: Choose Layout Strategy Combination}

For each dimension, provide reasoning and make a choice:
\begin{enumerate}[nosep,leftmargin=*]
\item \textbf{Content Arrangement:} linear-flow, grid-based, asymmetric, centered-focus, masonry, split-screen, sidebar-content, magazine-layout
\item \textbf{Component Grouping:} functional-clusters, visual-zones, priority-based, workflow-aligned, data-centric
\item \textbf{Space Allocation:} equal-distribution, primary-focus, golden-ratio, thirds-rule, flexible-grid
\item \textbf{Content Density:} spacious, balanced, compact, variable
\item \textbf{Visual Flow:} top-down, z-pattern, f-pattern, circular, focal-center
\end{enumerate}

\textbf{STEP 2:} Describe each component's layout using natural language (position, size, relationships)

\textbf{STEP 3:} Describe overall layout picture

\textbf{Return JSON format:}
\begin{verbatim}
{"chosen_strategies": {"content_arrangement": {"reasoning": "...",
   "choice": "grid-based"}},
 "overall_layout_description": "Description of full layout",
 "component_layouts": [{"id": "search-form",
   "layout_narrative": "Position and size description",
   "visual_prominence": "primary"}]}
\end{verbatim}
\end{tcolorbox}
\caption{Prompt for designing component layouts based on design analysis.}
\label{fig:prompt-layout-design}
\end{figure*}

\begin{figure*}[t]
\begin{tcolorbox}[title={\textbf{Prompt: Page Framework Generation}}, colback=gray!5, colframe=black, colbacktitle=black, coltitle=white]
\small
You are a senior web developer. Analyze the provided design image and generate a complete HTML framework with header and footer that matches the visual style.

\textbf{Website Seed:} \{website\_seed\}

\textbf{Header Navigation Links:} \{header\_links\_json\}

\textbf{Footer Links:} \{footer\_links\_json\}

\textbf{Design Analysis Context:} \{design\_context\}

\textbf{Requirements:}
\begin{enumerate}[nosep,leftmargin=*]
\item ANALYZE THE DESIGN IMAGE to extract: visual style, color palette, typography, layout patterns, spacing
\item Create a complete HTML framework matching the design (reusable for all pages)
\item Only include header, footer, and main content area (id=``content'')
\item Header matching the design's header style with provided navigation links
\item Footer matching the design's footer style with provided footer links
\item Modern, semantic HTML5 structure
\end{enumerate}

\textbf{CSS Requirements:}
\begin{itemize}[nosep,leftmargin=*]
\item Extract exact colors from the design image
\item Match typography from the design
\item Replicate spacing and sizing
\item Create CSS variables for the design system
\end{itemize}

\textbf{CRITICAL:}
\begin{itemize}[nosep,leftmargin=*]
\item Use English only
\item Do NOT include interactive elements without corresponding links
\item SVG files are not allowed in the framework
\end{itemize}

\textbf{Return JSON format:}
\begin{verbatim}
{"framework_html": "HTML with header/footer",
 "framework_css": "CSS replicating the design"}
\end{verbatim}
\end{tcolorbox}
\caption{Prompt for generating page framework (header/footer) from design image.}
\label{fig:prompt-page-framework}
\end{figure*}

\begin{figure*}[t]
\begin{tcolorbox}[title={\textbf{Prompt: HTML Page Generation}}, colback=gray!5, colframe=black, colbacktitle=black, coltitle=white]
\small
You are a senior web developer. Generate the main content HTML for a \{website\_type\} website page with UI JavaScript.

\textbf{Page Information:} \{page\_design\_json\}

\textbf{Navigation Information:} \{page\_architecture\_json\}

\textbf{Framework HTML:} \{framework\_html\}

\textbf{Data Dictionary:} \{data\_dict\_json\}

\textbf{Page-Specific SDK Interfaces:} \{page\_interfaces\_json\}

\textbf{Requirements:}
\begin{enumerate}[nosep,leftmargin=*]
\item Generate ONLY content for \texttt{<main id="content">} section
\item Call interfaces as \texttt{WebsiteSDK.functionName()} - they are SYNCHRONOUS
\item Handle incoming\_params: Extract URL parameters this page expects
\item Implement outgoing\_connections: Navigate to other pages with correct parameters
\item Add data attributes: \texttt{data-populate}, \texttt{data-action}, \texttt{data-component}
\end{enumerate}

\textbf{UI JavaScript Requirements:}
\begin{enumerate}[nosep,leftmargin=*]
\item Initialize page when DOM is ready
\item Extract URL parameters for incoming\_params
\item Call SDK methods based on \texttt{data-populate} attributes
\item Set up event listeners based on \texttt{data-action} attributes
\item Implement navigation with correct parameters
\item Always call \texttt{WebsiteSDK.methodName()} directly (no method extraction)
\end{enumerate}

\textbf{CRITICAL:} Call SDK interfaces with positional arguments only. Use only relative .html URLs for internal navigation.

\textbf{Return:} \texttt{\{"html\_content": "Complete HTML page with UI JavaScript"\}}
\end{tcolorbox}
\caption{Prompt for generating HTML pages with integrated UI JavaScript.}
\label{fig:prompt-html-generation}
\end{figure*}

\begin{figure*}[t]
\begin{tcolorbox}[title={\textbf{Prompt: CSS Page Generation}}, colback=gray!5, colframe=black, colbacktitle=black, coltitle=white]
\small
You are a senior web developer. Generate CSS styles for the page based on its HTML structure.

\textbf{Page Design:} \{page\_design\_json\}

\textbf{Page Layout:} \{page\_layout\_json\}

\textbf{Design Analysis:} \{design\_analysis\_json\}

\textbf{Framework CSS (build upon this):} \{framework\_css\}

\textbf{Generated HTML (style this content):} \{html\_content\}

\textbf{Requirements:}
\begin{enumerate}[nosep,leftmargin=*]
\item Include complete framework CSS - no abbreviations
\item Style the content area and page-specific components
\item Follow the design analysis color scheme and typography
\item Implement the layout specifications (grid, spacing, etc.)
\item Ensure responsive design with proper breakpoints
\item Use CSS variables defined in framework CSS
\item Add hover states and transitions for interactive elements
\item Use modern CSS features (flexbox, grid, custom properties)
\end{enumerate}

\textbf{CRITICAL:} Put this at the VERY TOP of css\_content:

\texttt{[hidden] \{ display: none !important; \}}

\textbf{Return:} \texttt{\{"css\_content": "Complete CSS including framework and page-specific styles"\}}
\end{tcolorbox}
\caption{Prompt for generating CSS styles based on HTML structure and design analysis.}
\label{fig:prompt-css-generation}
\end{figure*}

\begin{figure*}[t]
\begin{tcolorbox}[title={\textbf{Prompt: Data Generation}}, colback=gray!5, colframe=black, colbacktitle=black, coltitle=white]
\small
You are a data generator specializing in realistic website data. Generate comprehensive, realistic data based on the EXACT data dictionary specifications.

\textbf{Website Seed:} \{website\_seed\}

\textbf{User Tasks Context:} \{tasks\_json\}

\textbf{Data Dictionary Structure:} \{data\_types\_info\_json\}

\textbf{CRITICAL CONSTRAINTS:}
\begin{enumerate}[nosep,leftmargin=*]
\item \textbf{Use data\_type\_name as JSON key}: Use the exact value from ``data\_type\_name'' field
\item \textbf{Use EXACT field names}: Only fields defined in fields dictionary
\item \textbf{Follow field types}: string, number, boolean, array, datetime as specified
\item \textbf{Intelligent Volume Decision}: Based on generation\_type:
  \begin{itemize}[nosep,leftmargin=*]
  \item ``many'': Generate substantial amount approaching max\_items
  \item ``few'': Generate small representative set (20-30\% of max\_items)
  \end{itemize}
\item \textbf{No extra fields}: Do NOT add fields not in the dictionary
\end{enumerate}

\textbf{IMAGE URL REQUIREMENTS:} Use ONLY real, working image services:
\begin{itemize}[nosep,leftmargin=*]
\item Unsplash: \texttt{https://images.unsplash.com/photo-[ID]?w=800\&h=600}
\item Picsum: \texttt{https://picsum.photos/800/600?random=[1-1000]}
\end{itemize}

\textbf{DATA QUALITY:} Generate realistic, diverse content appropriate for the website seed. Ensure data relationships are logical and consistent.

\textbf{Return JSON format:}
\begin{verbatim}
{"static_data": {"products": [{"field1": "value"}],
  "categories": [{"id": "cat_1", "name": "Category"}]}}
\end{verbatim}
\end{tcolorbox}
\caption{Prompt for generating realistic website data based on data models.}
\label{fig:prompt-data-generation}
\end{figure*}

\begin{figure*}[t]
\begin{tcolorbox}[title={\textbf{Prompt: Backend Implementation Generation}}, colback=gray!5, colframe=black, colbacktitle=black, coltitle=white]
\small
You are an expert JavaScript developer. Generate a complete business logic implementation.

\textbf{Website Seed:} \{website\_seed\}

\textbf{Tasks:} \{tasks\_json\}

\textbf{Data Models:} \{data\_models\_json\}

\textbf{Interfaces:} \{interfaces\_json\}

\textbf{REQUIREMENTS:}
\begin{enumerate}[nosep,leftmargin=*]
\item Implement ALL core interfaces specified
\item Add helper functions as needed (prefix with \_ for private)
\item Use localStorage for ALL data persistence (browser-compatible)
\item NO DOM operations, NO window/document references (except localStorage)
\item Must work in both browser and Node.js environments
\item All data must be JSON serializable for localStorage
\item Implement interfaces with positional arguments only
\end{enumerate}

\textbf{STRUCTURE:}
\begin{verbatim}
const localStorage = (function() { ... })(); // polyfill
class BusinessLogic {
  constructor() { this._initStorage(); }
  _initStorage() { /* init localStorage tables */ }
  _getFromStorage(key) { /* retrieve data */ }
  _saveToStorage(key, data) { /* persist data */ }
  addToCart(productId, quantity) { /* implementation */ }
}
module.exports = BusinessLogic;
\end{verbatim}

\textbf{Return:} \texttt{\{"code": "javascript code here"\}}
\end{tcolorbox}
\caption{Prompt for generating business logic implementation.}
\label{fig:prompt-backend-implementation}
\end{figure*}

\begin{figure*}[t]
\begin{tcolorbox}[title={\textbf{Prompt: Backend Test Generation}}, colback=gray!5, colframe=black, colbacktitle=black, coltitle=white]
\small
You are an expert test engineer. Generate flow-based integration tests for the business logic.

\textbf{Website Seed:} \{website\_seed\}

\textbf{Tasks:} \{tasks\_json\}

\textbf{Interfaces:} \{interfaces\_json\}

\textbf{Generated Data:} \{generated\_data\_json\}

\textbf{CRITICAL REQUIREMENTS:}
\begin{enumerate}[nosep,leftmargin=*]
\item Use Generated Data ONLY in setupTestData() for initial localStorage population
\item NEVER hardcode expected return values - always extract from actual API responses
\item Chain API calls properly: Call API, capture response, extract needed values for next calls
\item Test complete user flows, not individual functions
\item Focus on happy path (successful scenarios)
\item Must run in Node.js environment
\item Test ALL tasks provided
\end{enumerate}

\textbf{CORRECT Flow Testing Example:}
\begin{verbatim}
const addResult = this.logic.addToCart(userId, productId, 2);
const actualCartId = addResult.cartId;  // Extract from response
const cartData = this.logic.getCart(actualCartId);  // Use actual ID
this.assert(cartData.total > 0, 'Total should be positive');
\end{verbatim}

\textbf{Return:} \texttt{\{"code": "javascript test code"\}}
\end{tcolorbox}
\caption{Prompt for generating flow-based integration tests.}
\label{fig:prompt-backend-tests}
\end{figure*}

\begin{figure*}[t]
\begin{tcolorbox}[title={\textbf{Prompt: Evaluator Generation}}, colback=gray!5, colframe=black, colbacktitle=black, coltitle=white]
\small
You are a QA engineer. Create evaluators to check if users complete tasks successfully.

\textbf{Website Seed:} \{website\_seed\}

\textbf{Tasks to evaluate:} \{tasks\_json\}

\textbf{Cross-Page States Structure:} \{cross\_page\_states\_json\}

\textbf{Generated Data Structure:} \{data\_structure\_json\}

For each task, create an evaluator that:
\begin{itemize}[nosep,leftmargin=*]
\item Uses cross-page states stored in localStorage to determine completion
\item Uses data structure knowledge to create precise validation logic
\item References exact field names and data types from the data structure
\item Provides clear evaluation criteria and logic
\item Uses JavaScript logic to check task completion status
\end{itemize}

\textbf{Guidelines:}
\begin{itemize}[nosep,leftmargin=*]
\item Use localStorage.getItem() to access both cross-page states and static data
\item Parse JSON data when retrieving complex objects from localStorage
\item Check for null/undefined values before accessing object properties
\item Use realistic validation logic based on the actual data structure
\end{itemize}

\textbf{Return JSON format:}
\begin{verbatim}
{"evaluators": [{"task_id": "task_1", "name": "Evaluator Name",
  "description": "What this evaluator checks",
  "localStorage_variables": ["selectedProductId", "products"],
  "evaluation_logic": "const products = JSON.parse(...); ..."}]}
\end{verbatim}
\end{tcolorbox}
\caption{Prompt for generating task completion evaluators.}
\label{fig:prompt-evaluator-generation}
\end{figure*}

\begin{figure*}[t]
\begin{tcolorbox}[title={\textbf{Prompt: Instrumentation Analysis}}, colback=gray!5, colframe=black, colbacktitle=black, coltitle=white]
\small
You are analyzing JavaScript business logic to determine what instrumentation variables are needed to evaluate task completion.

\textbf{TASKS TO EVALUATE:} \{tasks\_json\}

\textbf{CURRENT BUSINESS LOGIC:} \{code\_snippet\}

\textbf{EXISTING LOCALSTORAGE VARIABLES:} \{existing\_storage\_vars\_json\}

\textbf{DATA STORAGE KEYS:} \{storage\_keys\_json\}

\textbf{ANALYSIS REQUIREMENTS:}
For each task, determine:
\begin{enumerate}[nosep,leftmargin=*]
\item What operations must occur for the task to be considered complete?
\item Can we use existing localStorage variables to determine completion?
\item If NOT, what new instrumentation variables are needed?
\end{enumerate}

\textbf{INSTRUMENTATION GUIDELINES:}
\begin{itemize}[nosep,leftmargin=*]
\item Only add variables if existing localStorage is insufficient
\item Use naming convention: taskN\_actionDescription (e.g., task1\_searchCompleted)
\item Specify which function should set the variable and under what condition
\item Be conservative - only add instrumentation if truly necessary
\end{itemize}

\textbf{Return JSON:}
\begin{verbatim}
{"requirements": [{"task_id": "task_1",
  "needs_instrumentation": true,
  "required_variables": [{"variable_name": "task1_searchCompleted",
    "set_in_function": "searchNeighborhoods",
    "set_condition": "After successful search"}]}]}
\end{verbatim}
\end{tcolorbox}
\caption{Prompt for analyzing instrumentation requirements for task tracking.}
\label{fig:prompt-instrumentation-analysis}
\end{figure*}

\begin{figure*}[t]
\begin{tcolorbox}[title={\textbf{Prompt: Instrumentation Code Generation}}, colback=gray!5, colframe=black, colbacktitle=black, coltitle=white]
\small
You are adding instrumentation variables to JavaScript business logic for task completion tracking.

\textbf{ORIGINAL CODE:} \{original\_code\}

\textbf{INSTRUMENTATION SPECIFICATIONS:} \{instrumentation\_specs\_json\}

\textbf{INSTRUCTIONS:}
For each instrumentation variable:
\begin{enumerate}[nosep,leftmargin=*]
\item Find the specified function in the code
\item Add localStorage.setItem() call at the appropriate location based on set\_condition
\item Wrap instrumentation code in try-catch to ensure non-invasive behavior
\item Use the exact variable\_name and value\_to\_set from specifications
\end{enumerate}

\textbf{CRITICAL REQUIREMENTS:}
\begin{itemize}[nosep,leftmargin=*]
\item DO NOT change any original functionality
\item DO NOT modify function signatures or return values
\item Instrumentation code must be wrapped in try-catch
\item Only add localStorage.setItem() calls as specified
\item Preserve all existing code structure and comments
\item Place instrumentation BEFORE the return statement
\end{itemize}

\textbf{Return:} Complete instrumented business\_logic.js code
\end{tcolorbox}
\caption{Prompt for generating instrumented code with tracking variables.}
\label{fig:prompt-instrumentation-generation}
\end{figure*}

\begin{figure*}[t]
\begin{tcolorbox}[title={\textbf{Prompt: Instrumentation Evaluator Generation}}, colback=gray!5, colframe=black, colbacktitle=black, coltitle=white]
\small
You are generating evaluators to check if users completed tasks successfully.

\textbf{TASKS:} \{tasks\_json\}

\textbf{INSTRUMENTATION VARIABLES AVAILABLE:} \{var\_mapping\_json\}

\textbf{BUSINESS LOGIC IMPLEMENTATION:} \{business\_logic\_code\}

\textbf{WEBSITE DATA:} \{website\_data\_json\}

\textbf{INSTRUCTIONS:}
For each task, create an evaluator based on the instrumentation plan:

\textbf{Case 1: Tasks with needs\_instrumentation=true}
\begin{itemize}[nosep,leftmargin=*]
\item Use the instrumentation\_variables specific to that task
\item Validate the variable values match expected values
\end{itemize}

\textbf{Case 2: Tasks with needs\_instrumentation=false}
\begin{itemize}[nosep,leftmargin=*]
\item Use the existing\_variables to infer task completion
\item Check the ACTUAL data structure from the business logic implementation
\end{itemize}

All evaluators must:
\begin{itemize}[nosep,leftmargin=*]
\item Check if the variables exist in localStorage
\item Use the EXACT data structure from the business logic implementation
\item Return true if the task is completed, false otherwise
\end{itemize}

\textbf{Return JSON:}
\begin{verbatim}
{"evaluators": [{"task_id": "task_1", "name": "...",
  "localStorage_variables": ["var1", "var2"],
  "evaluation_logic": "// JavaScript returning boolean"}]}
\end{verbatim}
\end{tcolorbox}
\caption{Prompt for generating evaluators with instrumentation support.}
\label{fig:prompt-instrumentation-evaluator}
\end{figure*}

\end{document}